\documentclass[conference]{IEEEtran}
\usepackage{times}

\usepackage[numbers]{natbib}
\usepackage{multicol}
\usepackage[bookmarks=true]{hyperref}
\usepackage{xcolor}
\usepackage{amsmath}
\usepackage{amssymb}
\usepackage{tcolorbox}
\usepackage{booktabs}
\usepackage{multirow}
\usepackage{graphicx}
\usepackage{tabularx}
\usepackage{algorithm}
\usepackage{algpseudocode}
\usepackage{algorithmicx}
\usepackage{wrapfig}
\usepackage{subcaption}
\usepackage{mathtools}
\usepackage{bbm}
\usepackage{xcolor}
\usepackage{float}
\usepackage{placeins}

\usepackage{fancyhdr}

\usepackage{siunitx}
\begin{document}

\title{Action Flow Matching\\for Continual Robot Learning}

\author{\authorblockN{Alejandro Murillo-González and Lantao Liu}
\authorblockA{Indiana University--Bloomington\\
\texttt{\{almuri, lantao\}@iu.edu}}
}

\maketitle

\begin{abstract}

Continual learning in robotics seeks systems that can constantly adapt to changing environments and tasks, mirroring human adaptability. 
A key challenge is refining dynamics models, essential for planning and control, while addressing issues such as safe adaptation, catastrophic forgetting, outlier management, data efficiency, and balancing exploration with exploitation ---all within task and onboard resource constraints. 
Towards this goal, we introduce a generative framework leveraging flow matching for online robot dynamics model alignment.
Rather than executing actions based on a misaligned model, our approach refines planned actions to better match with those the robot would take if its model was well aligned. 
We find that by transforming the actions themselves rather than exploring with a misaligned model ---as is traditionally done--- the robot collects informative data more efficiently, thereby accelerating learning. 
Moreover, we validate that the method can handle an evolving and possibly imperfect model while reducing, if desired, the dependency on replay buffers or legacy model snapshots. 
We validate our approach using two platforms: an unmanned ground vehicle and a quadrotor. 
The results highlight the method’s adaptability and efficiency, with a record 34.2\% higher task success rate, demonstrating its potential towards enabling continual robot learning. Code: \url{https://github.com/AlejandroMllo/action_flow_matching}.

\end{abstract}

\IEEEpeerreviewmaketitle

\vspace{10pt}
\section{Introduction}

\begin{figure*}
    \centering
    \includegraphics[width=0.8\linewidth]{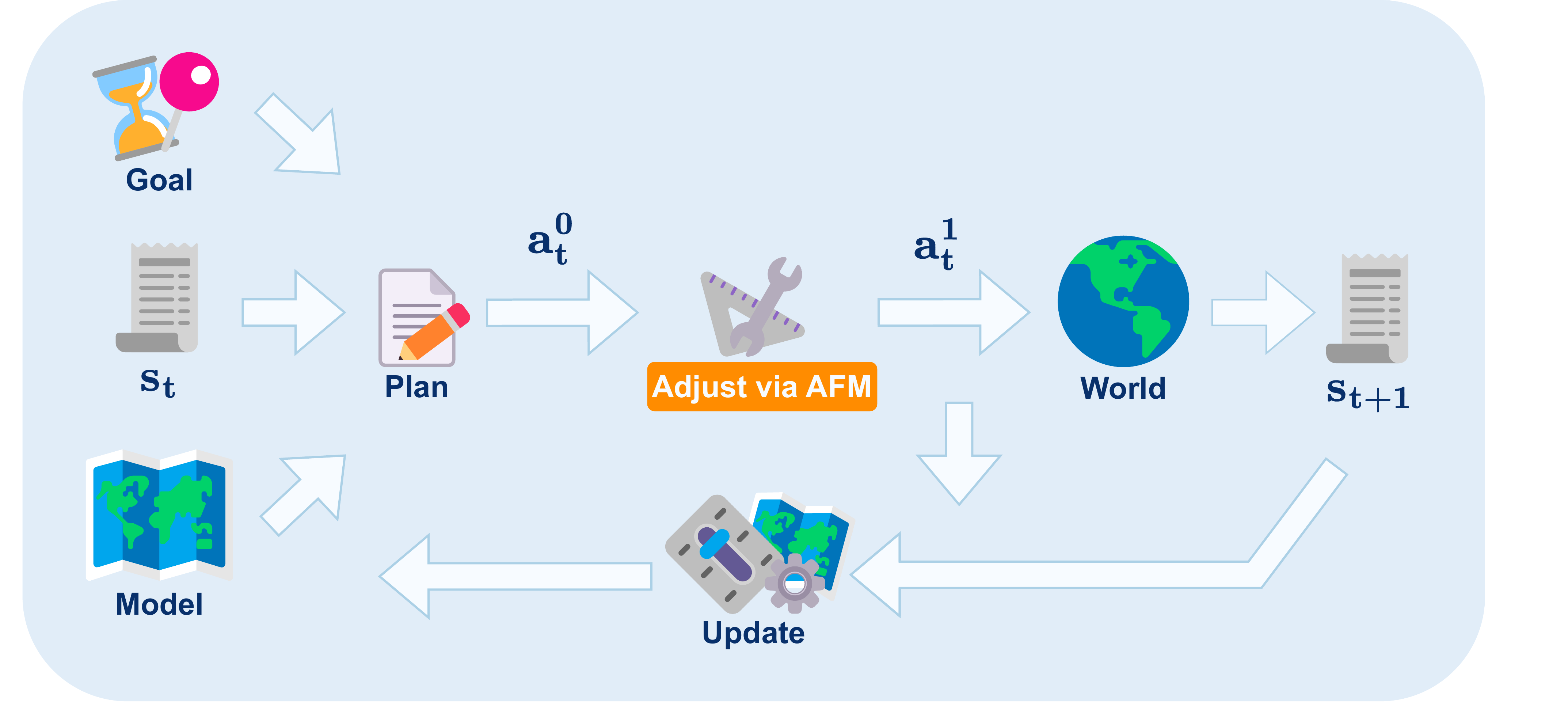}
    
    \caption{\small Action Flow Matching for efficient continual learning. The planner uses the latest version of the dynamics model 
    and the desired goal 
    to select the next action $\mathbf{a}_t^0$. AFM transforms $\mathbf{a}_t^0$ into $\mathbf{a}_t^1$ whenever the misalignment between the model and the environment is unacceptable. With $\mathbf{a}_t^1$, the agent is expected to execute more informative actions leading to faster model re-alignment.}
    \label{fig:overview}
\end{figure*}

Humans exhibit a remarkable ability to adapt in dynamic and uncertain environments, excelling at compensating for mismatches between expectations and actual outcomes. For example, when a driver encounters unexpectedly icy roads, they instinctively adjust their steering and braking actions to align with the anticipated altered dynamics caused by the changed road condition. This ability to plan effectively in the face of discrepancies inspires the robotics community to tackle a pressing challenge: enabling robots to operate robustly with potentially misaligned or limited dynamics models, where \textit{misalignment} can emerge from simplifying assumptions, new or unaccounted factors, incomplete or inaccurate data and parameter uncertainty. This capability is crucial for long-term autonomy, where unanticipated changes in environmental dynamics, such as reduced traction or actuator degradation, can render pre-designed, pre-learned or pre-calibrated models ineffective.

The dynamics model of a robot encapsulates how the robot predicts the outcome of its actions based on its state and control inputs. Accurate dynamics models form the foundation for planning and control, dictating how a robot predicts the outcomes of its actions to achieve desired goals. Techniques such as Linear Quadratic Regulator (LQR) \cite{kalman1960contributions, anderson2007optimal} and Model Predictive Control (MPC) \cite{richalet1978model, cutler1983dynamic, seborg2016process} rely critically on the fidelity of these models to generate and evaluate control policies. Without a robust and adaptable dynamics model, even advanced controllers falter, particularly in realistic scenarios where the robot must operate in diverse or evolving circumstances. For instance, modeling the effects of reduced traction on a slippery surface or compensating for actuator wear are essential for maintaining safe and effective robot operation. Dynamics models thus represent the link between the robot's perception, prediction, and control, enabling informed decision-making under uncertainty.

Although the potential for robots to improve their dynamics models through experiential learning is compelling, achieving this safely and effectively remains a significant challenge. Frequent model updates risk overfitting to transient phenomena or noise, leading to catastrophic forgetting of relevant previously acquired knowledge \cite{de2021continual}. Additionally, outliers and spurious correlations in data can compromise the model, undermining performance and safety. Conversely, overly conservative updating may fail to capture legitimate shifts in dynamics \cite{lesort2020continual}. 
Lifelong learning \cite{silver2013lifelong, thrun1998lifelong, khetarpal2022towards} introduces additional complexities, as exploration must be conducted efficiently within the context of ongoing task execution, rather than solely focusing on model improvement. The need to reconcile these trade-offs is further complicated by the diversity of environments a robot may encounter, each presenting unique and evolving dynamics. 

In this paper, we propose a novel method for continual learning of robot dynamics models that 
efficiently leverages the robot's latest experiences. Our approach departs from the conventional approach in current Model-Based Reinforcement Learning (MBRL) \cite{sutton2018reinforcement} where 
exploration is mostly guided via plans obtained with the latest version of the model \cite{chua2018deep, hansen2023td}. Specifically, we propose Action Flow Matching (AFM) to transform planned actions into those that more closely reflect the robot's actual \textit{intent}—defined as the action the robot would have selected if an aligned dynamics model was available.

Specifically, with a misaligned model, the planned actions are often suboptimal --or even totally wrong. 
This happens because the model fails to precisely account for the actual environmental and dynamical factors influencing the robot. However, the selected action gives us information about planned or desirable future states. 
AFM leverages such signals and learns a transformation that adjusts the initially planned actions, seeking to correct the discrepancies introduced by model inaccuracies, thus increasing the likelihood of reaching the planned state. In other words, the realigned actions can be viewed as the output of a calibration process correcting for the imperfect understanding of system dynamics, 
ensuring that the executed actions are more effective and informative for subsequent learning and control. Figure \ref{fig:overview} illustrates AFM’s role while updating a robot's dynamics model in an online and non-episodic manner.
 
Our key contributions can be summarized as follows:

\begin{itemize}
    \item \textbf{Efficient Dynamics Space Reduction:} 
    The initial model can be leveraged to \textit{reduce the state-action space to a dynamically feasible region}, which would otherwise be inefficient to capture through random trajectory generation. 
    Since at this stage we are not assuming access to a ground-truth model, we can only obtain from the initial model the reduction of the space to an approximate subset of feasible regions. The intuition behind this idea is demonstrated in Figure \ref{fig:feasible_dynamics_space}.
    \item \textbf{Representing the Dynamics Regime without Ground-Truth Models:} 
     By focusing on \textit{state evolution patterns}, we can learn to represent different \textit{dynamics regimes} without needing an accurate ground-truth model. Here it is important to highlight a crucial distinction—learning state ``evolution patterns" means merely identifying recurring patterns of state changes. This is unrelated to learning the underlying ``system dynamics" which are unavailable as long as the misalignment persists: we will capture the dynamics model via continual model updates.
    \item \textbf{Intent Mapping for Accelerated Learning:} Leveraging flow matching within the reduced dynamics space and the dynamics regime representations allows us to \textit{produce guiding signals} 
    that the agent can exploit. This, in turn, accelerates learning thanks to the collection of more informative transition information via transformed actions. 
\end{itemize}

\section{Related Work}

This section reviews related work, starting with continual dynamics learning for real-time system modeling, followed by online methods for robotic adaptation. We also briefly touch on batch-based approaches for robot adaptation, which, although not continual, provide valuable insights for enhancing robot performance and versatility.

\subsection{Continual Robot Dynamics Learning}

Learning a robot's dynamics online to fine tune or correct previous models has usually relied on strong inductive biases, which helps reduce the learning complexity but leads to task- or embodiment-specific methods. For example, \citet{jiahao2023online} learn online the residual dynamics of a drone following a reference trajectory. In this setting, the drone's mass changes as it operates, for instance because it needs to carry a load not considered in its original dynamics. Their work builds upon KNODE-MPC, which are knowledge-based neural ordinary differential equation methods, used to ``augment a model obtained from first principles" \cite{chee2022knode}. To turn KNODE-MPC into an online method in \cite{jiahao2023online}, the authors relied on a set of architecturally-equivalent models fine tuned sequentially and in parallel to the drone's operation. The robot then switches to the most up-to-date weights. However, to train the models, they require a replay buffer that relies on a window of past transition information, which requires additional memory requirements and selecting an additional hyperparameter (replay buffer window size). 

\citet{meier2016towards} learns the inverse dynamics error for a robotic arm. They focus on getting the residual terms for the model of a KUKA lightweight arm in simulation and real-world evaluation. They leverage Gaussian processes with an smoothed update rule to achieve controlled parameter updates, thus avoiding large parameter jumps detrimental for robustness. \citet{nguyen2011incremental} find a sparse representation of the data using a test of linear independence to get a fixed-size subset of the data that will be used for training. Then, the subset of data is used with an online kernel regression method to obtain the robot's model. However, note that this approach requires keeping both a database of the observed data and the varying subset of training data, which might be impractical for lifelong learning. The method is shown to work with a 7 degree-of-freedom Barrett WAM arm.

\citet{sun2021online} focus on dynamics learning in the context of legged locomotion. They conduct evaluations using a simulated biped and a real-world quadruped. Their approach involves learning a time-varying, locally linear residual model along the robot's current trajectory, aiming to correct the prediction errors of the controller's model. In particular, they use ridge regression to find the residual weights and biases of the nominal model. Initially, they rely on the nominal model and only start learning the residual parameter after $k$ timesteps, where $k$ is a new hyperparameter that helps them decide when enough data has been collected to be able to learn a good set of coefficients. $k$ also serves as the maximum size of the replay buffer, thus as new data arrives, old data is sequentially discarded to encourage learning only based on the latest transition dynamics.

\citet{jamone2013online} explore kinematics learning for a simulated humanoid working with tools with different shapes and sizes, which induces a non-negligible context switch. The context is recovered by identifying the transition discrepancy from similar inputs, which signals the presence of a different confounding factor (different tool in this case) that modifies the true robot's dynamics. To model the transitions they learn a multi-valued function, that allows for multiple potential solutions to be associated with the same query input. Nonetheless, they rely on the Infinite Mixture of Linear Experts algorithm, thus constraining the modeled relationships to collections of local linear models.

\subsection{Online Learning for Robot Adaptation}

\citet{pastor2011online} achieve online robotic arm movement adaptation by associating experienced sensor traces with stereotypical movements. The traces are leveraged to generate an implicit model that can be used to track differences in what the robot ``feels" for future similar movements. The model then enables planning reactive behaviors to compensate or adapt to unforeseen events. In contrast to our work, this work's adaptive behavior results from the predictive controller correcting for discrepancies based on the models learned a-priori, while we focus on modeling the new dynamics.

\citet{thor2019error} develop an online error-based learning method for frequency adaptation of central pattern generators (CPGs) to produce periodic motion patterns, which can be leveraged for robot locomotion. In this setting, learning happens using a dual integral learner that relies on a simple objective function that does not need to explicitly consider the system's dynamics. Although this method adds supporting evidence to error-guided learning for control adaptation, it is encapsulated to CPGs, which might not be relevant to controlling certain robotic platforms.

\citet{hagras2001prototyping, hagras2002online} achieve a lifelong learning approach for online controller learning and calibration. They focus on agricultural mobile robots that adapt their navigation behaviors to the dynamic and unstructured environment. Specifically, they learn online the membership functions for a hierarchical fuzzy genetic system. Additionally, each behavior is implemented as a simple fuzzy logic controller, thus limiting the number of inputs and outputs. Fuzzy coordination will then allow behaviors to activate concurrently at varying levels, providing smoother control than switching methods. 

Recently, \citet{nagabandi2018deep} propose MOLe which achieves continual learning in simulation environments by applying online stochastic gradient descent (SGD) to update the model parameters, while concurrently meta-learning a mixture model to represent parameters across different tasks. 
\citet{chen2020adaptive} do trajectory tracking for flexible-joint robots. The controller's network consists of a regressor and output layer, designed to model the robot's linear-in-parameter dynamics. The parameters are updated --guided by the trajectory tracking error-- on separate time scales: fast adaptation of output weights via Lyapunov-based laws and slower internal weight updates through backpropagation, inspired by singular perturbation theory.

\subsection{Batch-based Learning for Robot Adaptation}

The Rapid Motor Adaptation (RMA) family of algorithms has emerged to enable a robot to rapidly adapt to environmental challenges/particularities represented via a latent vector that augments the input of a policy learned offline, in simulation and using privileged information (e.g., ground truth sensor readings and environment information) \cite{kumar2021rma, kumar2022adapting, qi2023hand, liang2024rapid}. Although they have shown good performance, online learning is not possible given the need for multi-staged training, where the latent representation is first learned together with the model-free base policy, followed by learning an adaptation module that learns to predict the latent representation using past state-action data. \citet{lee2020context} propose Context-aware Dynamics Models (CaDM) for handling diverse dynamics. The approach involves learning a context encoder alongside forward and backward dynamics models, with both models conditioned on a latent vector that encodes dynamics regime-specific information. This work highlights the need for continual dynamics learning, as it is hard to exhaustively learn a robot's transition model and the challenge is exacerbated by the possibility of ambiguous state representations due to non-observed state variables (e.g., broken or missing sensors).

\section{Preliminaries}

We begin with an introduction to model-based planning and its reliance on predictive dynamics models for effective decision-making. Then, we introduce flow matching, a generative framework for transforming random variables from a source to a target distribution.

\subsection{Model-Based Planning and Control}

Let \( \mathcal{S} \subseteq \mathbb{R}^n \) be the state space representing all possible configurations of the system. Let \( \mathcal{A} \subseteq \mathbb{R}^m \) be the action space representing all possible control actions that influence the system's dynamics. Model-based planning leverages a dynamics model \( f: \mathcal{S} \times \mathcal{A} \to \mathcal{S} \) to predict the state evolution. 
This predictive capability allows planners to optimize a sequence of actions \( \mathbf{a} \in \mathcal{A} \) from a state $\mathbf{s} \in \mathcal{S}$ by minimizing a cost function \( \mathcal{J}(\mathbf{s}, \mathbf{a}) \) that balances the multiple objectives of the robot, for example, goal $\mathcal{G}$ completion, obstacle avoidance, and resource efficiency:
\begin{equation}
\mathbf{a}^* = \arg\min_{\mathbf{a} \in \mathcal{A}^H} \mathcal{J}(\mathbf{s}_t, \mathbf{a}; H).
\end{equation}
where the trajectory cost \( \mathcal{J} \) is defined as:
\begin{equation}
\mathcal{J}(\mathbf{s}_t, \mathbf{a}; H) = \sum_{h=0}^{H-1} \ell(\mathbf{s}_{t+h}, \mathbf{a}_{t+h}) + \Phi(\mathbf{s}_{t+H}).
\end{equation}
Here, \( \mathbf{a} = [\mathbf{a}_t, \mathbf{a}_{t+1}, \dots, \mathbf{a}_{t+H-1}] \) represents the sequence of actions over the planning horizon \( H \), while \( \ell \) and \( \Phi \) denote the running cost at each timestep and the terminal cost, respectively. The sequence of states \( \{\mathbf{s}_t, \mathbf{s}_{t+1}, \dots, \mathbf{s}_{t+H}\} \) is determined recursively using the dynamics model \( f \) as:
\begin{equation}
\mathbf{s}_{t+1} = f(\mathbf{s}_t, \mathbf{a}_t),
\end{equation}
where \( \mathbf{s}_t \in \mathcal{S} \) represents the state of the system at time \( t \), and \( \mathbf{a}_t \in \mathcal{A}\) is the action at \( t \). 

In this setting, the dynamics model $f$ serves as the mathematical representation of the constraints governing the evolution of the system's state. That is, the feasibility of any proposed sequence \( \mathbf{a} \) is dictated by the system's dynamics.

Mathematically, this turns the optimization into a constrained sequential decision-making problem:
\begin{align} \label{eq:mpc-opti-problem}
&\mathbf{a}^* = \arg\min_{\mathbf{a}} \mathcal{J}(\mathbf{s}_t, \mathbf{a}, H), \\ 
\mathtt{subject~to} \quad &\mathbf{s}_{t+h+1} = f(\mathbf{s}_{t+h}, \mathbf{a}_{t+h}), \nonumber\\
&\mathbf{a}_{t+h} \in \mathcal{A}, \quad \forall h \in [0, H-1]. \nonumber
\end{align}

This formulation emphasizes that \( f \), the dynamics model, governs admissible trajectories by encoding critical physical constraints such as non-linearities, dynamics limits, and uncertainties. Any inaccuracies in \( f \) shift the feasible solution space, resulting in plans \( \{\mathbf{s}_t\} \) that may violate real-world constraints or be suboptimal when executed. Consequently, maintaining an aligned dynamics model is essential for effective planning and control.

We solve the optimization problem in Eq. \eqref{eq:mpc-opti-problem} using sampling-based methods, which are well-suited for high-dimensional and non-linear dynamics. These methods optimize action sequences by sampling trajectories, evaluating their costs, and selecting those that minimize the objective \( \mathcal{J} \). The differences among this family of methods lie in how the candidate distribution is selected and updated at decision time, with examples including Model Predictive Path Integral (MPPI) \cite{gandhi2021robust, mohamed2022autonomous} and the Cross Entropy Method (CEM) \cite{botev2013cross}.

The critical role of the dynamics model highlights the need for mitigating the impact of model misalignment, particularly in complex, unstructured environments where exhaustively anticipating all changes in transition dynamics is impractical or even infeasible. AFM, introduced later in this work, addresses this by transforming planned actions into those that better align with the agent’s true \textit{intention}, compensating for errors caused by misalignment between the true and modeled dynamics.

\subsection{Flow Matching (FM)}

In the following overview we adopt the notation and description of FM by \citet{lipman2024flowmatchingguidecode}: FM is a framework for generative modeling that constructs bijective, deterministic flows through a learned velocity field, designed to interpolate between a source $p$ and target $q$ distribution. Formally, given a training dataset with samples from a target distribution \( q \subseteq \mathbb{R}^d \), the goal is to train a neural network parameterized as a time-dependent velocity field \( u^\phi_\tau : [0,1] \times \mathbb{R}^d \to \mathbb{R}^d \) such that the associated flow transforms the source distribution \( p \) into \( q \). This is achieved by constructing a time-continuous probability path \( \{ p_\tau \}_{\tau=0}^1 \), satisfying \( p_0 = p \) and \( p_1 = q \). Procedurally, the method can be decomposed as follows:

\subsubsection{Constructing the Flow}
The flow transformation \( \psi_\tau : \mathbb{R}^d \to \mathbb{R}^d \) is defined through the solution of an Ordinary Differential Equation (ODE) parameterized by the velocity field $u^\phi_\tau$:
\begin{equation}
\frac{\mathrm{d}\psi_\tau(x)}{\mathrm{d}\tau} = u^\phi_\tau(\psi_\tau(x)), \quad \psi_0(x) = x, \quad \tau \in [0,1].
\end{equation}
Given \( \psi_\tau \), the intermediate distribution \( p_\tau \) is generated by transforming samples \( X_0 \sim p \) through \( X_\tau = \psi_\tau(X_0) \).

\subsubsection{Designing the Probability Path}

This is used to interpolate samples from the source to the target distribution. Among the alternatives, we choose the {\em conditional optimal-transport path}, which defines the random variable $X_\tau \sim p_\tau$ as the linear combination of the source $X_0 \sim p$ and target $X_1 \sim q$ samples:
\begin{equation}
    X_\tau = \tau X_1 + (1 - \tau) X_0 \sim p_\tau.
\end{equation}

\subsubsection{Training Objective}
The core learning objective is to minimize the discrepancy between the learned velocity field \( u^\phi_\tau \) and the target velocity \( u_\tau \), defined via a regression loss:
\begin{equation} \label{eq:loss_fm}
\mathcal{L}_{\text{FM}}(\phi) = \mathbb{E}_{\tau \sim U[0,1], X_\tau \sim p_\tau} \big\| u^\phi_\tau(X_\tau) - u_\tau(X_\tau) \big\|^2.
\end{equation}

However, in general, obtaining the ground-truth $u_\tau$ is not tractable, as it corresponds to a complex entity that governs the joint transformation between two high-dimensional distributions \cite{lipman2024flowmatchingguidecode}. 

Therefore, we employ a conditional formulation of the loss:
\begin{align} \label{eq:cfm_loss}
    &\mathcal{L}_{\text{CFM}}(\phi) = \\ \nonumber
    & \mathbb{E}_{\tau \sim U[0,1], X_0 \sim p, X_1 \sim q} \big\| u^\phi_\tau ( (1-\tau)X_0 + \tau X_1) - \frac{X_1 - X_0}{1-\tau} \big\|^2,
\end{align}
whose key property is that it shares the same gradient as the nominal flow matching loss:  
\begin{equation}
    \nabla_\phi \mathcal{L}_{\text{FM}}(\phi) = \nabla_\phi \mathcal{L}_{\text{CFM}}(\phi),
\end{equation}
thus enabling learning a valid velocity field $u_\tau^\phi$ without the need for the ground-truth $u_\tau$.

\subsubsection{Sampling from the Model}
After training, new samples $X_1 \sim q$ are drawn by solving the ODE over the interval \( \tau \in [0,1] \) for an input \( X_0 \sim p \):
\begin{equation}
X_1 = \psi_1(X_0).    
\end{equation}
Specific to our work, we obtain $X_1$ using the learned velocity fields, which are integrated via an explicit midpoint ODE solver \cite{suli2003introduction}.

\section{Method: Action Flow Matching} \label{sec:method}

\begin{figure*}[h]
    \centering
    \includegraphics[width=0.92\linewidth]{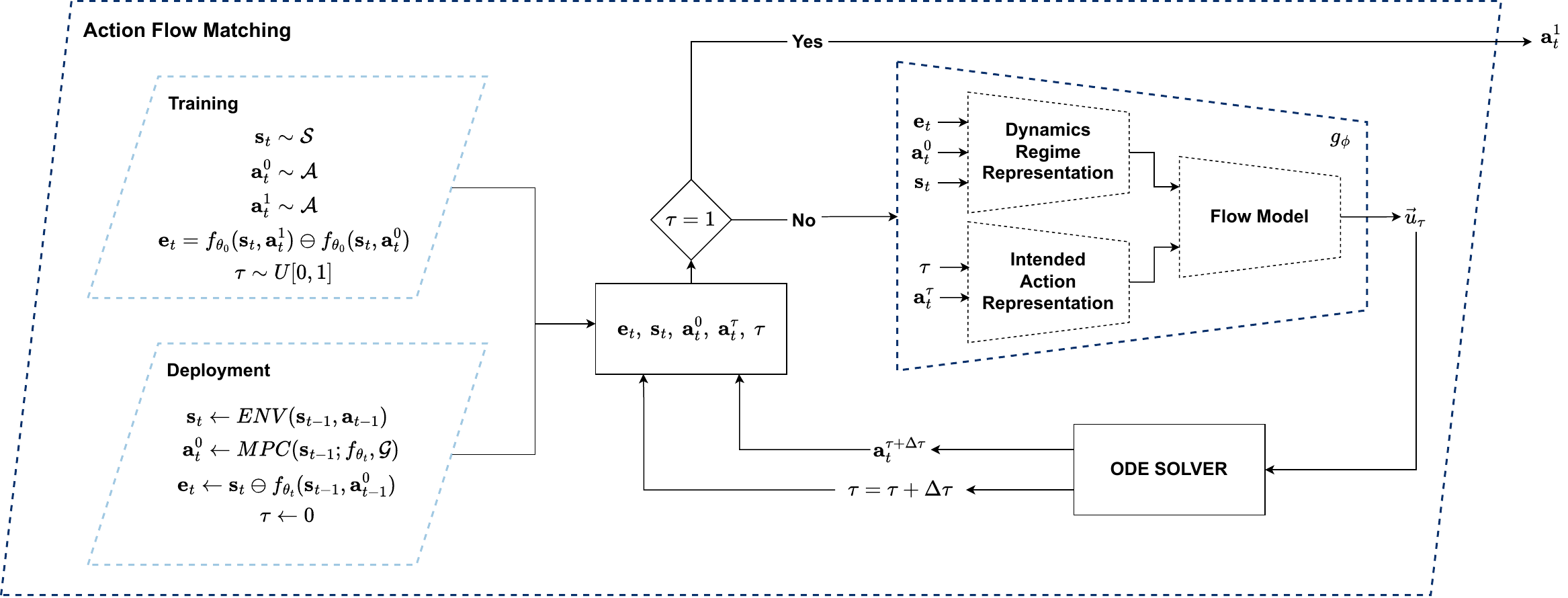}
    \caption{\small AFM Pipeline. During training we leverage counterfactual-inspired transitions by simulating the state evolution with one randomly sampled action but registering a different random action as the cause. At deployment time, the agent tries to identify the current dynamics regime and based on that transforms the planned actions $\mathbf{a}_t^0$ to a corrected or intended action $\mathbf{a}_t^1$ that AFM considers has a higher chance of accomplishing the desired state transition.}
    \label{fig:afm}
\end{figure*}

We introduce Action Flow Matching (AFM), 
a method towards continual learning in robotics built on a set of key insights, to leverage misaligned models for faster deployment-time model alignment. In this context, misalignment can arise from simplifying assumptions, new or unaccounted latent factors, incomplete or inaccurate training data and parameter uncertainty. 

Our identified insights include:
(1) \textit{efficient dynamics space reduction} using the initially available model;
(2) \textit{dynamics regime representation learning without ground-truth models} by focusing on state change evolution patterns; and
(3) \textit{intent mapping for accelerated learning} by using a generative model to produce guiding signals based on information about the intent of the robot and the current dynamics regime, instead of solely relying on the possibly misaligned model.

\subsection{Formal Overview}

Given the available dynamics model \( f_{\theta_t} \) at time $t$, with weights $\theta_t$ updated every step using the previous state transitions, the agent predicts the system's future states over a horizon \( H \) using planned actions \( \mathbf{a}_t^0 \), as follows:
\begin{equation}
    \mathbf{s}_{t+1} = f_{\theta_t}(\mathbf{s}_t, \mathbf{a}_t^0), \quad \text{for } t = 0, \ldots, H-1.
\end{equation}
Here, \( \mathbf{a}_t^0 \) is chosen to minimize a trajectory cost \( \mathcal{J}(\mathbf{s}_t, \mathbf{a}_t, H) \), which evaluates deviations from desired outcomes and control effort. However, when \( f_{\theta_t} \) significantly deviates from the true environment dynamics \( f \), the predicted states \( \mathbf{s}_{t+1} \) fail to accurately reflect the system's behavior, leading to suboptimal plans and limiting both learning and control performance.

To address these discrepancies, we deviate from the conventional approach of only relying on the imperfect dynamics \( f_\theta \) to gather experience with which the model can be improved, since complex systems can make it challenging and time consuming. Instead, we opt to add a more direct and efficient strategy: \textit{transforming the action \( \mathbf{a}_t^0 \) itself}.

The essence of our proposed framework can be likened to optometry, where instead of attempting to correct the biological intricacies of a myopic person's eyes—an inherently complex and delicate system—we place corrective lenses that {\em adjust the incoming light path}, ensuring the perceived image is properly aligned with reality. Similarly, rather than solely modifying the intricate and often imperfect system dynamics, we introduce an additional transformation step that {\em compensates for discrepancies}, ensuring that the executed actions are better aligned with the intended outcomes. This additional step in our case is AFM, whose overview is presented in Figure \ref{fig:afm}. 

Formally, the transformed action \( \mathbf{a}_t^1 \) aims to reduce the deviation between the \textit{planned} next state:
\begin{equation} \label{eq:desired_next_state}
    \mathbf{s}_{t+1}^* = f_{\theta_t} (\mathbf{s}_t, \mathbf{a}_t^0),
\end{equation} 
and the state the robot would actually reach if $\mathbf{a}_t^0$ is executed (\textit{realized} next state):
\begin{equation} \label{eq:realized_next_state}
    \mathbf{s}_{t+1} = f(\mathbf{s}_t, \mathbf{a}_t^0),
\end{equation}
that is,
\begin{align} \label{eq:transformation-objective}
    \| f(\mathbf{s}_t, \mathbf{a}_t^1) -  \mathbf{s}_{t+1}^* \|_2  \leq \| f(\mathbf{s}_t, \mathbf{a}_t^0) - \mathbf{s}_{t+1}^* \|_2.
\end{align}

It is important to highlight that the constraint in Eq. \eqref{eq:transformation-objective} aims to reduce the deviation, instead of minimizing it. The main drivers for this problem formulation are twofold: 
\begin{itemize}
    \item \textit{Limited access to environment dynamics.} We cannot assume access to the current environment dynamics \( f \).
    Minimizing \( \| \mathbf{s}_{t+1} - \mathbf{s}_{t+1}^* \|_2 \) would require full knowledge of \( f \), which remains partially unknown and is the exact target we seek to learn. As such, attempting to directly minimize this quantity would render the learning of the AFM model \( g_\phi \) both intractable and redundant. 
    \item \textit{Enhancing data informativeness and model alignment.} By targeting the more tractable constraint in Eq. \eqref{eq:transformation-objective}, the agent is encouraged to collect \textit{more} informative state evolution data. This not only facilitates faster alignment of \( f_{\theta_t} \), 
    but also improves planning efficacy by ensuring that actions \( \mathbf{a}_t^1 \) more closely correspond to the desired outcomes. These features are especially advantageous for lifelong learning algorithms \cite{silver2013lifelong, thrun1998lifelong}.
\end{itemize}

\subsection{Learning to Map Plans to Intentions}

To enable efficient continual learning of evolving dynamics, we focus on generating informative exploratory trajectories. This addresses the challenge posed by misaligned dynamics models $f_{\theta_t}$, where planned actions $\mathbf{a}_t^0$ may fail to achieve the desired outcomes, resulting in data collection inefficiencies in regions of the state space with low informational value. 

To mitigate this, we propose the AFM model $g_\phi$, designed to transform the distribution of planned actions $\mathbf{a}_t^0$ into actions $\mathbf{a}_t^1$ that align more closely with the agent's intended actions. Here, the term \emph{intended actions} refers to those that would have been chosen had the agent used a well-aligned dynamics model. Consequently, AFM guides the agent to select actions that promote data collection in regions of higher utility, thereby expediting model alignment.

The key contribution of AFM stems from its novel training strategy, which enables the mapping of planned actions under a potentially misaligned dynamics model to actions that more closely achieve the intended outcomes. Standard approaches to flow matching would require datasets consisting of samples from the source and target distributions. Specifically, AFM would necessitate a dataset comprising planned actions $\mathbf{a}_t^0$ (source distribution) 
and intended actions $\mathbf{a}_t^1$ (target distribution). 
However, creating such datasets is infeasible due to the challenging hurdle of comprehensively identifying the causes and consequences of model misalignments. Additionally, direct access to \( f \) or the desired action distribution is unavailable (both are what we are actually trying to learn). Therefore, the diversity and extent of possible source distributions 
introduce the following two critical challenges:

\begin{itemize}
    \item[\textbf{Q1}] How can we generate diverse datasets that simulate dynamics model misalignments while relying solely on the initial dynamics model $f_{\theta_0}$? This model, learned from data representing a specific dynamics regime (e.g., flat office carpet), may not generalize to other regimes (e.g., icy sidewalk).
    \item[\textbf{Q2}] How can we identify and characterize the current dynamics regime using only observed state evolution data, thereby informing AFM of environmental characteristics and the predictive reliability of the current model $f_{\theta_t}$?
\end{itemize}

\begin{wrapfigure}{l}{0.4\linewidth}
    \centering
    \includegraphics[width=\linewidth]{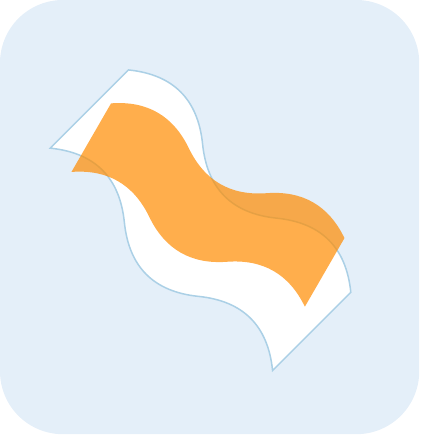}
    \caption{\small Feasible Dynamics Space Intuition. \textit{Blue}: State-Action space. \textit{White}: Dynamically feasible region. \textit{Orange}: Approximation of the feasible region using the dynamics model.} 
    \label{fig:feasible_dynamics_space}
\end{wrapfigure}

To address \textbf{Q1}, one naive approach would involve randomly generating tuples $(\mathbf{s}_t, \mathbf{a}_t^1, \mathbf{s}_{t+1})$ alongside corresponding planned actions $\mathbf{a}_t^0$ and using them to train the flow matching model. However, this would inefficiently cover the entire state-action space, disregarding the physical constraints introduced by the agent's embodiment. Instead, we leverage the initial dynamics model $f_{\theta_0}$ to approximate the \emph{dynamically feasible region}, depicted in Figure~\ref{fig:feasible_dynamics_space}. This allows us to constrain the exploration space, ensuring focus on regions of practical relevance. Mathematically, we define the feasible space as follows:
\begin{equation}
    f_{\theta_0}(\mathbf{s}, \mathbf{a}) \in \mathbb{F}, \quad \forall \mathbf{s} \in \mathcal{S}, ~ \forall \mathbf{a} \in \mathcal{A},
\end{equation}
where $\mathbb{F} \subseteq \mathbb{R}^{n}$ represents the approximated feasible dynamics region derived from $f_{\theta_0}$.

Therefore, to populate the state evolution dataset $\mathcal{E}$ used to train $g_\phi$, we simulate how actions influence state transitions under varying dynamics regimes. Specifically, given an initial state $\mathbf{s}_t \sim \mathcal{S}$, a \textit{planned} action $\mathbf{a}_t^0 \sim \mathcal{A}$, and an \textit{intended} action $\mathbf{a}_t^1 \sim \mathcal{A}$, we compute the next state $\mathbf{s}_{t+1}$ under the intended action:
\begin{equation}
    \mathbf{s}_{t+1} = f_{\theta_0}(\mathbf{s}_t, \mathbf{a}_t^1),
\end{equation}
but \textit{register the transition as if it resulted from the planned action}:
\begin{equation}
    \mathbf{s}_{t+1} \sim f_{\theta_0}(\mathbf{s}_t, \mathbf{a}_t^0).
\end{equation}
Consequently, the state evolution dataset $\mathcal{E}$ is constructed as follows:
\begin{equation}
    \mathcal{E} \coloneqq \left\{ \left( \mathbf{s}_i, \mathbf{a}_i^0, f_{\theta_0}( \mathbf{s}_i, \mathbf{a}_i^1 ) \right) : \mathbf{s}_i \sim \mathcal{S},~\mathbf{a}_i^0,\mathbf{a}_i^1 \sim \mathcal{A} \right\}_{i=0}^N
\end{equation}
We find this counterfactual-inspired misaligned dynamics simulation generates diverse and useful source and target distributions for AFM.

To address \textbf{Q2}, we propose a learned representation $\mathbf{Z}_{\mathcal{D}}$ that encodes the current dynamics regime. This representation is produced by a model that receives the current state $\mathbf{s}_t$, the planned action $\mathbf{a}_t^0$, and the vector difference (which we denote $\ominus$) between the planned state transition $\mathbf{s}_{t+1}^*$ (Eq. \ref{eq:desired_next_state}) and the realized state transition $\mathbf{s}_{t+1}$ (Eq. \ref{eq:realized_next_state}). Note, however, that we need to consider how to compute \(\mathbf{e}_t\) in two distinct scenarios:

\begin{itemize}
    \item During the dataset generation phase, an error value \(\mathbf{e}_t\) is calculated as the difference between the predicted state transitions when using two actions, \(\mathbf{a}_t^1\) (the action that results in the observed state transition) and \(\mathbf{a}_t^0\) (the planned action):
    \begin{equation}
        \mathbf{e}_t = f_{\theta_0}(\mathbf{s}_t, \mathbf{a}_t^1) \ominus f_{\theta_0}(\mathbf{s}_t, \mathbf{a}_t^0).
    \end{equation}
    \item During operation in the environment, the error is instead defined as the difference between the observed next state, \(\mathbf{s}_{t+1}\), and the predicted next state based on the planned (untransformed) action \(\mathbf{a}_t^0\):
    \begin{align}
        \mathbf{e}_t = \mathbf{s}_{t+1} \ominus f_{\theta_t}(\mathbf{s}_t, \mathbf{a}_t^0).
    \end{align}
\end{itemize}

In addition to the dynamics regime representation model, \(\mathtt{E}_{\mathbf{Z}_\mathcal{D}} : \mathcal{S} \times \mathcal{A} \times \mathcal{S}_{\ominus} \rightarrow \mathbb{R}^{\text{dim}(\mathbf{Z}_\mathcal{D})}\), the AFM model \(g_\phi\) includes a transformed action encoder, \(\mathtt{E}_{\mathbf{Z}_\mathcal{T}} : [0, 1] \times \tilde{ \mathcal{A} } \rightarrow \mathbb{R}^{\text{dim}(\mathbf{Z}_\mathcal{T})}\), where $\tilde{ \mathcal{A} } \subseteq \mathbb{R}^m $ symbolizes the action transformations space. The latent representations \(\mathbf{Z}_\mathcal{D}\) and \(\mathbf{Z}_\mathcal{T}\) are passed through a flow model, \(\mathtt{FM} : \mathbb{R}^{\text{dim}(\mathbf{Z}_\mathcal{D})} \times \mathbb{R}^{\text{dim}(\mathbf{Z}_\mathcal{T})} \rightarrow \mathbb{R}^{\text{dim}(\mathcal{S})}\), responsible for predicting the velocity field $\vec{u}_\tau \in \mathbb{R}^{dim(\mathbf{a}_t)}$:
\begin{align}
    \vec{u}_\tau & = g_\phi(\mathbf{s}_t, \mathbf{a}_t^0, \mathbf{e}_t, \tau, \mathbf{a}_t^\tau), \quad \tau \in [0, 1], \\
     & = \mathtt{FM} \Big( 
     \mathtt{E}_{\mathbf{Z}_\mathcal{D}}(\mathbf{s}_t, \mathbf{a}_t^0, \mathbf{e}_t; \phi_{\mathbf{Z}_\mathcal{D}}), 
     \mathtt{E}_{\mathbf{Z}_\mathcal{T}}(\tau, \mathbf{a}_t^\tau; \phi_{\mathbf{Z}_\mathcal{T}}); 
     \phi_{\mathtt{FM}} 
     \Big), \nonumber
\end{align}
where \(\phi = [\phi_{\mathbf{Z}_\mathcal{D}}; \phi_{\mathbf{Z}_\mathcal{T}}; \phi_{\mathtt{FM}}]\) represents the learnable parameters of the model. Using the collected dataset $\mathcal{E}$, learning the model $g_\phi$ involves optimizing the parameters $\phi$ to minimize the conditional flow matching loss $\mathcal{L}_{\text{CFM}}$ defined in Eq.~\eqref{eq:cfm_loss}. 

Finally, to compute \(\mathbf{a}_t^1\) from \(\mathbf{a}_t^0\), the velocity field \(\vec{u}_\tau\) is used by an explicit midpoint ODE solver \cite{suli2003introduction}. This solver takes incremental steps, transforming \(\mathbf{a}_t^\tau\) into \(\mathbf{a}_t^{\tau + \Delta \tau}\) (starting from $\tau=0$), iteratively refining the action toward \(\mathbf{a}_t^1\). 

\subsection{Continual Learning with AFM}

Once the AFM model \(g_\phi\) is trained, it can be leveraged for accelerated continual dynamics learning. This process involves using decision-time planning to select actions that explore regions of the space most informative for reducing model uncertainty, thus accelerating adaptation to new dynamics regimes. 

A crucial aspect is determining when to rely solely on the learned model \(f_{\theta_t}\) and when to utilize the AFM model to mitigate potential model misalignments. This decision process is formalized as follows. At each iteration, we record the predicted state transition \(\mathbf{s}_{\text{pred}} = f_{\theta_t}(\mathbf{s}_{t-1}, \mathbf{a}_{t-1}^0)\) and the realized next state \(\mathbf{s}_{\text{real}}\). The prediction error \(\mathbf{e}_t\) is computed as the \(L_2\) norm of the difference between the realized and predicted states:
\begin{equation}
    \mathbf{e}_t = 
\begin{cases} 
\mathbf{0}, & \text{if } \|\mathbf{s}_{\text{real}} - \mathbf{s}_{\text{pred}}\|_2 < \delta_\mathcal{M}, \\
\mathbf{s}_{\text{real}} - \mathbf{s}_{\text{pred}}, & \text{otherwise},
\end{cases}
\end{equation}
where \(\delta_\mathcal{M}\) is a predefined threshold for acceptable dynamics misalignment. From \(\mathbf{e}_t\), a binary model misalignment flag \(\mathcal{M}\) is generated:
\begin{equation}
    \mathcal{M} = \mathbbm{1} \left(\|\mathbf{e}_t\|_2 > 0 \right).
\end{equation}
The flag \(\mathcal{M}\) dictates whether action transformations are required. If \(\mathcal{M}\) transitions from \(\mathtt{True}\) to \(\mathtt{False}\), it implies that the current model \(f_{\theta_t}\) aligns with the dynamics, allowing actions to be executed without transformation. Conversely, a transition from \(\mathcal{M} = \mathtt{False}\) to \(\mathcal{M} = \mathtt{True}\) indicates a dynamics regime change. In this case, transformed actions should be executed to facilitate faster model re-alignment.

Upon detecting a regime change, a representation of the current dynamics \(\mathbf{Z}_{\mathcal{D}}\) is computed as:
\begin{equation}
    \mathbf{Z}_{\mathcal{D}} = \mathtt{E}_{\mathbf{Z}_\mathcal{D}}(\mathbf{s}_t, \mathbf{a}_t^0, \mathbf{e}_t; \phi_{\mathbf{Z}_\mathcal{D}}),
\end{equation}
where \(\mathtt{E}_{\mathbf{Z}_\mathcal{D}}\) encodes the relevant contextual information for \(g_\phi\). The AFM model then maps the planned action \(\mathbf{a}_t^0\) to a transformed action $\mathbf{a}_t^1$:
\begin{align}
    \mathbf{a}_t^1 = \mathtt{ODE\_SOLVER} \Big( &\mathtt{FM} \big( 
     \mathbf{Z}_{\mathcal{D}}, 
     \mathtt{E}_{\mathbf{Z}_\mathcal{T}}(\tau, \mathbf{a}_t^\tau; \phi_{\mathbf{Z}_\mathcal{T}}); 
     \phi_{\mathtt{FM}} 
     \big), \nonumber \\ & \quad \tau_s =0, \tau_e=1, step=\Delta \tau \Big)
\end{align}
 which is executed every time step until the model is realigned (flag \(\mathcal{M}\) transitions again).

Furthermore, at each time step, the tuple \(\mathcal{E}_{t-1} = (\mathbf{s}_{t-1}, \mathbf{a}_{t-1}, \mathbf{s}_t)\) is collected and used to update the model parameters \(\theta_{t-1}\) to \(\theta_t\). 
Note that our approach makes no assumptions about the specifics of \(f_{\theta}\), as it is agnostic to the architecture, training methodology, and other design choices of the dynamics model. Instead, our focus is on generating more informative actions that facilitate adapting \(f_{\theta}\), thereby enabling and accelerating continual learning.

\section{Results}

\begin{figure}[h!]
    \centering
    \includegraphics[width=\linewidth]{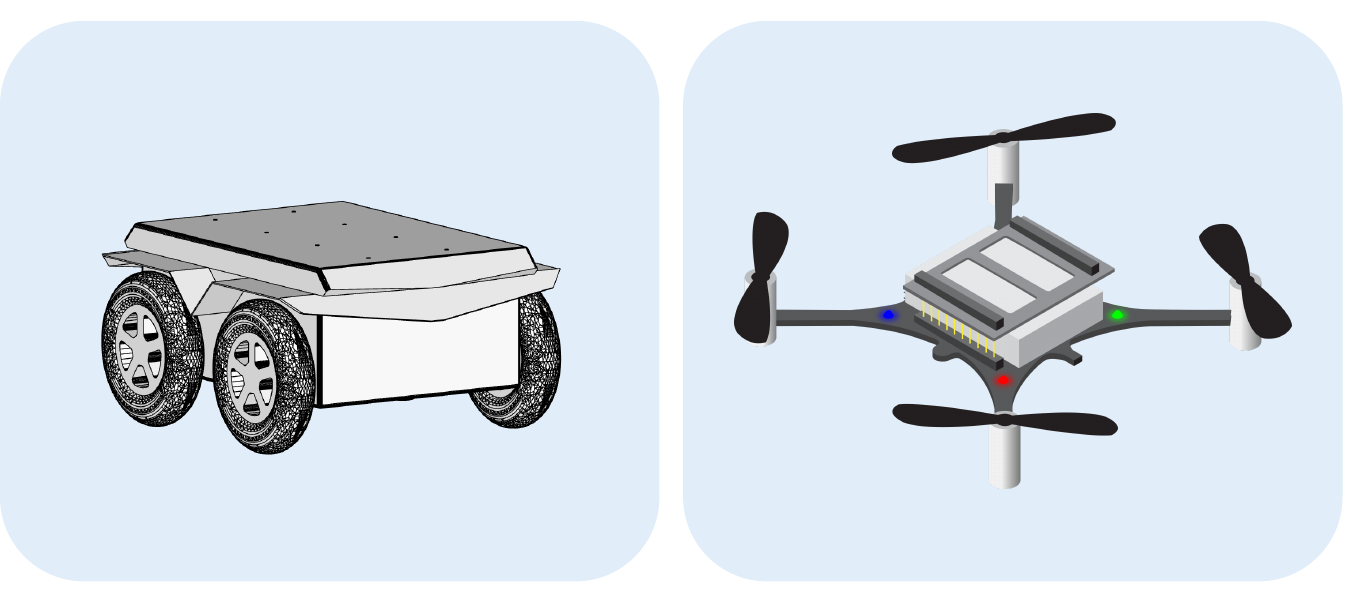}
    \caption{\small Platforms employed for validation of our method and comparison against the baselines in simulation. \textit{(Left)} Jackal UGV \cite{jackal_website}. \textit{(Right)} Crazyflie quadrotor \cite{bitcraze_crazyflie_simulation}.}
    \label{fig:platforms}
\end{figure}

We evaluate our method on two robotic platforms shown in Figure \ref{fig:platforms}: the Jackal UGV and the Crazyflie quadrotor. The experiments assess whether AFM enables faster transition model alignment under evolving world or embodiment conditions that affect the robot's dynamics. Adapting to these changes is crucial for successfully completing the tasks.

\begin{figure}
    \centering
    \includegraphics[width=0.49\linewidth]{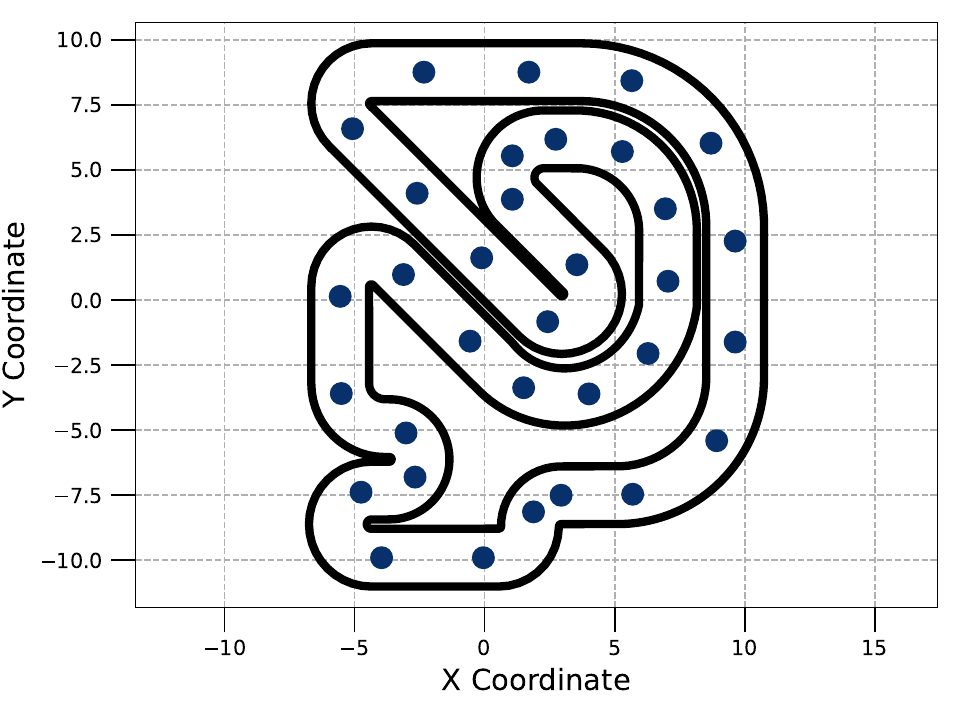}
    \includegraphics[width=0.49\linewidth]{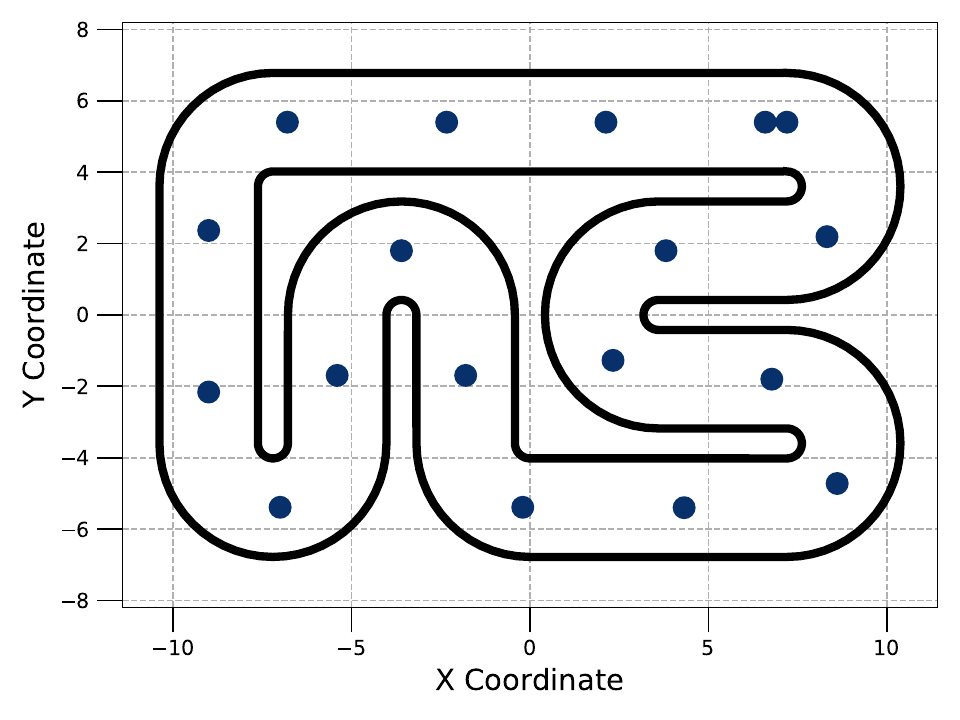}
    \caption{\small ETHZ Maps \cite{liniger2015optimization} where the online dynamics learning tasks took place. The UGV has to reach a series of sparse waypoints (blue dots) to complete the lap. At certain points the dynamics change and the robot needs to adapt to the new and unexpected world characteristics.}
    \label{fig:ugv-tasks-maps}
\end{figure}

\subsection{Baselines}

We are interested in comparing with methods that achieve dynamics model adaptation by learning from the continuous stream of robot observations during deployment\footnote{Approaches that focus on offline adaptation strategies \cite{kumar2021rma}, latent-space dynamics learning \cite{hansen2023td}, assume some structure over the task space \cite{cao2022provable, ruvolo2013ella}, or directly learn a policy \cite{stachowiczlifelong} fall outside the scope of this paper.}:

\begin{itemize}

    \item \textbf{stream-x PE} \cite{elsayed2024streaming}: This work introduces strategies to enable batch-based model-free reinforcement learning methods to operate in a streaming fashion. We adopt these ideas and integrate them into the probabilistic ensemble (PE) framework \cite{chua2018deep} to evaluate their performance in a model-based context. 
    We provide an introduction to the stream-x family of algorithms in Appendix \ref{appendix:stream-x}.
    \item \textbf{Online-KNODE-MPC} \cite{jiahao2023online}: This method utilizes Knowledge-based Neural Ordinary Differential Equations (KNODE) combined with transfer learning techniques to achieve continuous dynamics model alignment for a quadrotor. A significant limitation of this approach is its reliance on prior knowledge about the system dynamics, represented as a set of differential equations. This requirement can be impractical for systems with highly nonlinear or poorly understood dynamics. It also requires maintaining multiple past copies of the model, 
    which can become problematic when running on resource constrained platforms.  

    \item \textbf{Physics}: Physics-based models, commonly employed in model predictive control (MPC), serve as an essential baseline for comparison. These models rely on well-known principles of physics for each platform. For the UGV we use the Dubins \cite{dubins-dyns} model and for the Quadrotor we adopt the model from \cite{chee2022knode}. 
    They provide a strong benchmark for evaluating the benefits of adaptive dynamics models in tasks requiring high accuracy and adaptability.
\end{itemize}

\subsection{Navigating under Evolving Dynamics}

\begin{figure}
    \centering
    \includegraphics[width=\linewidth]{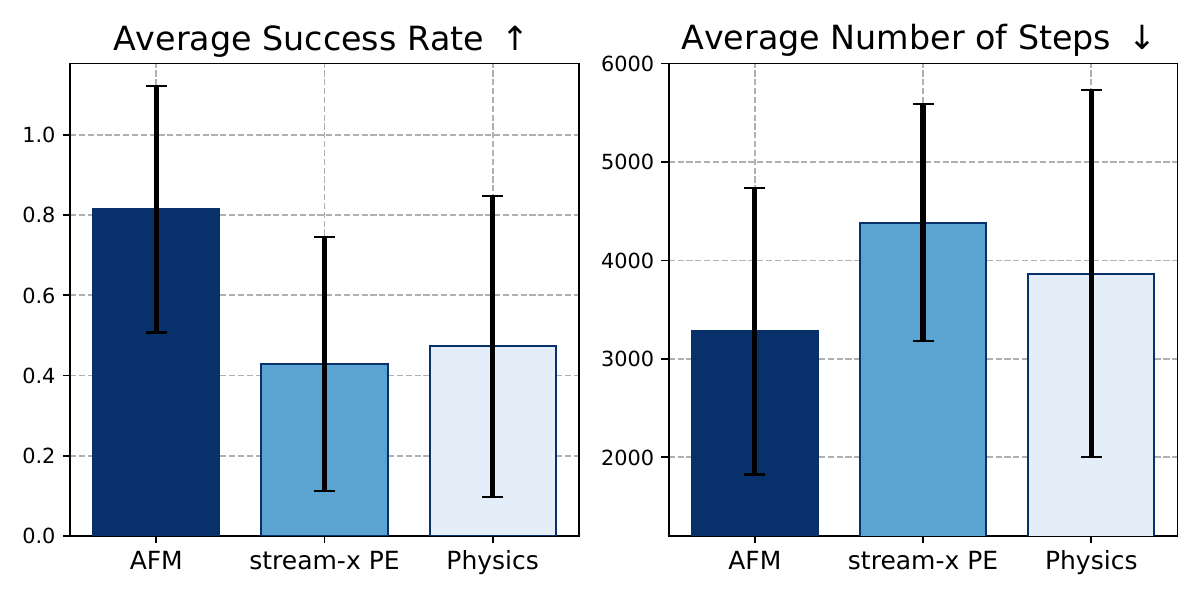}
    \caption{\small Aggregated results for online and non-episodic UGV dynamics learning. Mean $\pm$ standard deviation over five seeds.}
    \label{fig:ugv_aggregated_results}
\end{figure}

\begin{table*}[h!]
\centering
\caption{\small UGV Lifelong Dynamics Learning Results in Map 1 with $\delta_\mathcal{M} = 1$. Results averaged ($\pm$ standard deviation) over five seeds. Best results are shown in \textbf{bold}, while the least favorable results are highlighted in {\color{gray}gray}.}
\label{tab:results_summary_map1}
\begin{tabular}{cccccccc}
\toprule
\multicolumn{2}{c}{\textbf{Scenario}} & \multicolumn{6}{c}{\textbf{Method}} \\
\cmidrule(lr){1-2} \cmidrule(lr){3-8}
\multirow{{2}}{*}{\textbf{$v_\text{gain}$}} & \multirow{{2}}{*}{\textbf{$\omega_\text{gain}$}} & \multicolumn{2}{c}{\textbf{AFM (Ours)}} & \multicolumn{2}{c}{\textbf{stream-x PE}} & \multicolumn{2}{c}{\textbf{Physics}} \\
\cmidrule(lr){3-4} \cmidrule(lr){5-6} \cmidrule(lr){7-8}
 & & \textbf{Success Rate} $\uparrow$ & \textbf{Steps} $\downarrow$ & \textbf{Success Rate} $\uparrow$ & \textbf{Steps} $\downarrow$ & \textbf{Success Rate} $\uparrow$ & \textbf{Steps} $\downarrow$ \\
\midrule
-1.00 & 2.00 & \textbf{1.00 $\pm$ 0.00} & \textbf{2863.40 $\pm$ 169.90} & 0.22 $\pm$ 0.06 & {\color{gray}4999.0 $\pm$ 0.00} & {\color{gray}0.18 $\pm$ 0.00} & {\color{gray}4999.0 $\pm$ 0.00} \\
-2.50 & 1.00 & \textbf{1.00 $\pm$ 0.00} & \textbf{3037.60 $\pm$ 442.18} & 0.28 $\pm$ 0.09 & {\color{gray}4999.0 $\pm$ 0.00} & {\color{gray}0.18 $\pm$ 0.00} & {\color{gray}4999.0 $\pm$ 0.00} \\
2.50 & 0.05 & {\color{gray}0.18 $\pm$ 0.00} & {\color{gray}4999.0 $\pm$ 0.00} & 0.21 $\pm$ 0.02 & {\color{gray}4999.0 $\pm$ 0.00} & \textbf{0.32 $\pm$ 0.00} & {\color{gray}4999.0 $\pm$ 0.00} \\
1.00 & -1.00 & \textbf{1.00 $\pm$ 0.00} & 1825.60 $\pm$ 189.14 & {\color{gray}0.89 $\pm$ 0.21} & {\color{gray}3569.80 $\pm$ 769.58} & 1.00 $\pm$ 0.00 & \textbf{827.00 $\pm$ 10.28} \\
1.00 & -0.50 & 0.71 $\pm$ 0.27 & 4120.60 $\pm$ 1082.86 & {\color{gray}0.55 $\pm$ 0.23} & {\color{gray}4588.40 $\pm$ 821.20} & \textbf{1.00 $\pm$ 0.00} & \textbf{1673.00 $\pm$ 47.47} \\
-1.00 & 1.00 & \textbf{1.00 $\pm$ 0.00} & \textbf{3035.80 $\pm$ 277.85} & {\color{gray}0.18 $\pm$ 0.00} & {\color{gray}4999.0 $\pm$ 0.00} & {\color{gray}0.18 $\pm$ 0.00} & {\color{gray}4999.0 $\pm$ 0.00} \\
-0.50 & 1.00 & \textbf{1.00 $\pm$ 0.00} & \textbf{4625.00 $\pm$ 327.35} & {\color{gray}0.18 $\pm$ 0.00} & {\color{gray}4999.0 $\pm$ 0.00} & {\color{gray}0.18 $\pm$ 0.00} & {\color{gray}4999.0 $\pm$ 0.00} \\
2.00 & 2.00 & \textbf{1.00 $\pm$ 0.00} & 662.80 $\pm$ 31.38 & 1.00 $\pm$ 0.00 & {\color{gray}1913.60 $\pm$ 86.53} & 1.00 $\pm$ 0.00 & \textbf{472.80 $\pm$ 11.65} \\
-1.00 & -1.00 & \textbf{1.00 $\pm$ 0.00} & \textbf{3363.20 $\pm$ 345.08} & 0.44 $\pm$ 0.21 & {\color{gray}4999.0 $\pm$ 0.00} & {\color{gray}0.18 $\pm$ 0.00} & {\color{gray}4999.0 $\pm$ 0.00} \\
0.10 & -1.50 & 0.23 $\pm$ 0.03 & {\color{gray}4999.0 $\pm$ 0.00} & {\color{gray}0.18 $\pm$ 0.01} & {\color{gray}4999.0 $\pm$ 0.00} & \textbf{0.74 $\pm$ 0.00} & {\color{gray}4999.0 $\pm$ 0.00} \\
-0.50 & 0.50 & \textbf{0.76 $\pm$ 0.06} & {\color{gray}4999.0 $\pm$ 0.00} & {\color{gray}0.18 $\pm$ 0.00} & {\color{gray}4999.0 $\pm$ 0.00} & {\color{gray}0.18 $\pm$ 0.00} & {\color{gray}4999.0 $\pm$ 0.00} \\
\midrule
\bottomrule
\end{tabular}
\vspace{0.6cm}
\centering
\caption{\small UGV Lifelong Dynamics Learning Results in Map 2 with $\delta_\mathcal{M} = 1$. Results averaged ($\pm$ standard deviation) over five seeds. Best results are shown in \textbf{bold}, while the least favorable results are highlighted in {\color{gray}gray}.}
\label{tab:results_summary_map2}
\begin{tabular}{cccccccc}
\toprule
\multicolumn{2}{c}{\textbf{Scenario}} & \multicolumn{6}{c}{\textbf{Method}} \\
\cmidrule(lr){1-2} \cmidrule(lr){3-8}
\multirow{{2}}{*}{\textbf{$v_\text{gain}$}} & \multirow{{2}}{*}{\textbf{$\omega_\text{gain}$}} & \multicolumn{2}{c}{\textbf{AFM (Ours)}} & \multicolumn{2}{c}{\textbf{stream-x PE}} & \multicolumn{2}{c}{\textbf{Physics}} \\
\cmidrule(lr){3-4} \cmidrule(lr){5-6} \cmidrule(lr){7-8}
 & & \textbf{Success Rate} $\uparrow$ & \textbf{Steps} $\downarrow$ & \textbf{Success Rate} $\uparrow$ & \textbf{Steps} $\downarrow$ & \textbf{Success Rate} $\uparrow$ & \textbf{Steps} $\downarrow$ \\
\midrule
-1.00 & 2.00 & \textbf{1.00 $\pm$ 0.00} & \textbf{1929.00 $\pm$ 225.27} & 0.33 $\pm$ 0.26 & {\color{gray}4999.0 $\pm$ 0.00} & {\color{gray}0.16 $\pm$ 0.00} & {\color{gray}4999.0 $\pm$ 0.00} \\
-2.50 & 1.00 & \textbf{0.83 $\pm$ 0.34} & \textbf{3557.60 $\pm$ 744.91} & 0.55 $\pm$ 0.32 & {\color{gray}4999.0 $\pm$ 0.00} & {\color{gray}0.16 $\pm$ 0.00} & {\color{gray}4999.0 $\pm$ 0.00} \\
2.50 & 0.05 & {\color{gray}0.16 $\pm$ 0.00} & {\color{gray}4999.0 $\pm$ 0.00} & 0.19 $\pm$ 0.03 & {\color{gray}4999.0 $\pm$ 0.00} & \textbf{0.40 $\pm$ 0.04} & {\color{gray}4999.0 $\pm$ 0.00} \\
1.00 & -1.00 & \textbf{1.00 $\pm$ 0.00} & 1082.00 $\pm$ 96.05 & {\color{gray}0.97 $\pm$ 0.06} & {\color{gray}3461.80 $\pm$ 1270.92} & 1.00 $\pm$ 0.00 & \textbf{595.40 $\pm$ 11.64} \\
1.00 & -0.50 & \textbf{1.00 $\pm$ 0.00} & 1981.00 $\pm$ 526.56 & 1.00 $\pm$ 0.00 & {\color{gray}2278.00 $\pm$ 574.91} & 1.00 $\pm$ 0.00 & \textbf{1222.20 $\pm$ 305.80} \\
-1.00 & 1.00 & \textbf{1.00 $\pm$ 0.00} & \textbf{2516.40 $\pm$ 420.09} & 0.33 $\pm$ 0.27 & {\color{gray}4999.0 $\pm$ 0.00} & {\color{gray}0.16 $\pm$ 0.00} & {\color{gray}4999.0 $\pm$ 0.00} \\
-0.50 & 1.00 & \textbf{1.00 $\pm$ 0.00} & \textbf{3884.60 $\pm$ 342.32} & {\color{gray}0.16 $\pm$ 0.00} & {\color{gray}4999.0 $\pm$ 0.00} & {\color{gray}0.16 $\pm$ 0.00} & {\color{gray}4999.0 $\pm$ 0.00} \\
2.00 & 2.00 & \textbf{1.00 $\pm$ 0.00} & 353.00 $\pm$ 14.18 & 1.00 $\pm$ 0.00 & {\color{gray}653.40 $\pm$ 57.22} & 1.00 $\pm$ 0.00 & \textbf{276.20 $\pm$ 5.91} \\
-1.00 & -1.00 & \textbf{1.00 $\pm$ 0.00} & \textbf{2754.00 $\pm$ 423.29} & 0.29 $\pm$ 0.27 & {\color{gray}4999.0 $\pm$ 0.00} & {\color{gray}0.16 $\pm$ 0.00} & {\color{gray}4999.0 $\pm$ 0.00} \\
0.10 & -1.50 & 0.19 $\pm$ 0.03 & {\color{gray}4999.0 $\pm$ 0.00} & {\color{gray}0.16 $\pm$ 0.00} & {\color{gray}4999.0 $\pm$ 0.00} & \textbf{0.94 $\pm$ 0.08} & \textbf{4983.80 $\pm$ 15.32} \\
-0.50 & 0.50 & \textbf{0.86 $\pm$ 0.12} & \textbf{4915.60 $\pm$ 110.71} & {\color{gray}0.16 $\pm$ 0.00} & {\color{gray}4999.0 $\pm$ 0.00} & {\color{gray}0.16 $\pm$ 0.00} & {\color{gray}4999.0 $\pm$ 0.00} \\
\midrule
\bottomrule
\end{tabular}
\end{table*}

\textbf{Unmanned Ground Vehicle:} The Jackal UGV --shown in Figure \ref{fig:platforms} \textit{(left)}-- is a robust wheeled platform commonly used for research in ground vehicle dynamics and navigation. This platform allows us to rigorously evaluate the ability of adaptive dynamics models to handle complex interactions such as varying traction conditions and actuator degradation during long-term deployment. Its state $\mathbf{s}_t = [x_t, y_t, \theta_t]^T \in \mathcal{S} \subseteq \mathbb{R}^3$ corresponds to the pose and heading of the robot. The control actions $\mathbf{a}_t = [v_t, \omega_t] \in \mathcal{A} \subseteq \mathbb{R}^2$ are the linear $v_t \in [-1, 1]$ and angular $\omega_t \in [-\pi/2, \pi/2]$ velocities.

\textbf{Environments and Tasks:} We conduct our experiments on two \textit{upscaled} ETHZ test tracks \cite{liniger2015optimization}. 
These tracks present a balanced mix of operational challenges, combining smooth, straightforward sections with tighter curves and more complex geometry. The task requires the robot to follow a set of sparse waypoints, shown as blue dots in Figure \ref{fig:ugv-tasks-maps}. A key challenge arises when, after reaching 15\% of the waypoints, the environment dynamics undergo a significant shift, necessitating rapid adaptation for the robot to successfully complete the task. To further evaluate the system, the dynamics revert to the nominal configuration after 80\% of the waypoints are achieved, introducing another adaptation requirement. The change in the dynamics regime is achieved via gains intervening the simulator's control variables. Notably, the robot is unaware of the timing or occurrence of these dynamic transitions, relying solely on its internal dynamics regime representation to predict and adapt effectively. The agent is allotted 5,000 steps to complete each task; exceeding this limit results in the task being deemed incomplete. We evaluate the models according to their success rate and average number of steps to reach all waypoints.

\textbf{Dynamics Model:} As the initial or base model we employ an ensemble of probabilistic neural networks \cite{chua2018deep}. The architecture explicitly models both epistemic and aleatoric uncertainties, enabling it to handle complex dynamics with inherent stochasticity. Despite its success in various settings, the PE baseline traditionally assumes offline or batch training, limiting its applicability to streaming or continuously evolving environments. To train $f_{\theta_0}$ we use the Negative Log-Likelihood (NLL) loss with respect to the predicted mean $\hat{\mu}$ and variance $\hat{\sigma}^2$: $\mathcal{L}(\theta) = - \log \hat{\sigma}^2 ||\hat{\mu}, y||_2 + \log \hat{\sigma}^2$ using 50.000 randomly generated training samples from the Dubins physics-based model \cite{dubins-dyns}, 256 batch size and the Adam optimizer with the default parameters in PyTorch 2.5.1 for 32 epochs or until the loss change between epochs was below 1e-3. During deployment we update the model's weights $\theta_{t-1}$ to get the latest dynamics model $f_{\theta_t}$ using the same loss function and optimizer, but only leveraging the latest transition tuple (i.e., replay buffer size is 1). We use an ensemble of size 5 with two hidden linear layers, each with 200 units, and Leaky ReLU nonlinearities in between. For planning with $f_{\theta_t}$ we use MPPI \cite{gandhi2021robust} with population size 1000, $\gamma=0.9,~ \sigma=0.4,~\text{and}~\beta=0.6$.

\textbf{AFM Model:} The action encoder maps the input action and a time-dependent parameter through two linear layers with 64 hidden units and ELU activation in between. The dynamics encoder processes the current state, action, and next state using a similar structure. The flow network combines both encodings through three linear layers (with 128, 64 and 64 hidden units, respectively) to predict the velocity field. The model is trained using the Adam optimizer (learning rate 0.01) with an MSE loss and a batch size of 256. The training data is generated using the procedure introduced in Section \ref{sec:method} in batches of 2048 samples for 75.000 iterations. Note that the same AFM model (weights and architecture) was used for all scenarios, supporting its usefulness across tasks without needing to assume any kind of task structure.

\textbf{Results:} Figure \ref{fig:ugv_aggregated_results} summarizes the results across tasks and environments, demonstrating that AFM consistently outperforms all baselines. Specifically, as shown in Tables \ref{tab:results_summary_map1} and \ref{tab:results_summary_map2}, AFM achieves an average success rate of \textbf{81.4\%}, which is \textbf{34.2\%} higher than the best-performing baseline that has 47.2\% success rate. Moreover, our approach requires \textbf{15.1\%} fewer steps on average (3277.95 vs 3865.24), confirming its superior efficiency.

A detailed breakdown of results reveals that our method exhibits robustness across diverse environments, maintaining a high success rate with minimal variance (0.09 for AFM vs 0.14 for the best baseline). In contrast, baseline methods exhibit significant performance degradation in challenging scenarios, especially in environments requiring substantial velocity adaptation by the robot. In 15 out of 22 scenarios, the best-performing baseline exhausted the entire allotted time without successfully completing the task. In contrast, our method encountered this issue \textit{in only 5 out of 22} scenarios. This means AFM had a \textbf{66\%} improvement in the most challenging scenarios. Moreover, in none of these five cases did the other methods consistently complete the task, highlighting the overall difficulty of the task and the robustness of our approach in comparison.

\textit{A key distinction of our method is that it does not assume any inherent task structure}, making the adaptation process inherently more challenging. Unlike other approaches that leverage predefined inductive biases to facilitate learning \cite{cao2022provable, ruvolo2013ella}, our method must generalize effectively without relying on prior knowledge of task distributions. This constraint necessitates a more robust and flexible adaptation mechanism, capable of handling diverse and previously unseen scenarios with minimal assumptions about underlying task properties.

The observed improvements stem from our method’s ability to \textit{balance exploration and exploitation based on the misalignment level and correcting actions that seek more informative regions of the space}. Overall, these results highlight the efficacy of AFM in achieving both higher task success and greater sample efficiency. More importantly, they represent a significant step towards the ultimate goal of lifelong robot dynamics learning, where robots continuously adapt and refine their models over extended periods in diverse environments.

\subsection{Flying under Evolving Dynamics}

\textbf{Quadrotor:} The Crazyflie quadrotor --shown in Figure \ref{fig:platforms} \textit{(right)}-- is a lightweight, highly agile aerial robot that offers unique challenges due to its underactuated nature and sensitivity to aerodynamic and other exogenous effects.
Its state $\mathbf{s}_t = [x_t, y_t, z_t, \dot{x}_t, \dot{y}_t, \dot{z}_t, q_{x,t}, q_{y,t}, q_{z,t}, q_{w,t}, \omega_{x,t}, \omega_{y,t}, \omega_{z,t}]^T \in \mathcal{S} \subseteq \mathbb{R}^{13}$ includes the position, linear velocity, orientation represented as a quaternion $(q_{x,t}, q_{y,t}, q_{z,t}, q_{w,t})$, and angular velocity. The control actions $\mathbf{a}_t = [u_{1,t}, u_{2,t}]^T \in \mathcal{A} \subseteq \mathbb{R}^{4}$ consist of the total thrust $u_{1,t} \in \mathbb{R}$ and the moment vector $u_{2,t} \in \mathbb{R}^3$, which are functions of the individual motor forces.

\textbf{Environment and Task:} We follow the same settings and evaluation metrics as \cite{jiahao2023online}. The quadrotor needs to track circular trajectories with a 2 m radius at target speeds of 0.3, 0.8, 1.0, 1.2, and 1.7 m/s. Each simulation runs for 8 seconds at 500 Hz. To assess adaptiveness, the quadrotor's mass is altered mid-flight: reduced by 50\% at 2 seconds and increased to 133\% of the nominal mass at 5 seconds. Performance is assessed as the MSE between the reference trajectory and the state of the robot every time step.

\textbf{Dynamics Model:} To verify that AFM is agnostic to the underlying dynamics model, we leverage the author-provided weights and implementation of Online-KNODE-MPC \cite{jiahao2023online} as the initial model $f_{\theta_0}$. All hyperparmeters are left unchanged. 

\textbf{AFM Model:} The model is analogous to the one used for the UGV platform. The only difference is that we generate 15M training samples using the initial model and the procedure from Section \ref{sec:method} to learn $g_\phi$.

\begin{wrapfigure}{l}{0.45\linewidth}
    \centering
    \includegraphics[width=\linewidth]{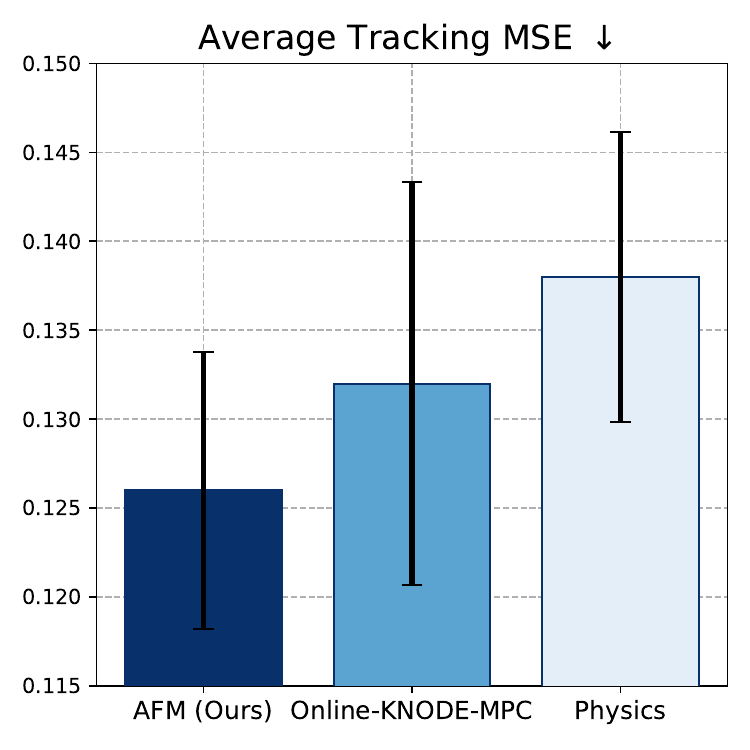}
    \caption{\small Quadrotor trajectory tracking error under evolving dynamics. Mean $\pm$ standard deviation over five experiments.}
    \label{fig:quad_aggregated_results}
\end{wrapfigure}

\textbf{Results:} Figure \ref{fig:quad_aggregated_results} shows that integrating AFM with the Online-KNODE-MPC quadrotor dynamics model reduces the average tracking error. Across baselines, AFM reduced the MSE by \textbf{6.6\%} on average, which is quiet significant as these feature expert-designed physics-based models that already enabled low tracking error provided the designer's assumptions remained valid. Since all settings of Online-KNODE-MPC remain unchanged except for its planned actions $\mathbf{a}_t^0$, which are refined by AFM, this improvement can be attributed to AFM’s corrections. By guiding the robot toward informative regions, AFM accelerates the update of the model $f_{\theta_t}$, enhancing overall performance.  

A closer analysis of the results reveals that this reduction in tracking error is most pronounced in scenarios with rapidly changing dynamics, where conventional MPC-based physics-only approaches struggle due to growing model inaccuracies. The ability of AFM to refine actions ensures that the quadrotor can better compensate for the multiple unmodeled disturbances and dynamic shifts it experienced during task execution.

\begin{table*}[h!]
\centering
\caption{\small Ablation Results. UGV Lifelong Dynamics Learning in Map 1 with $\delta_\mathcal{M} = 1$. Results averaged ($\pm$ standard deviation) over five seeds. Best results are shown in \textbf{bold}, while the least favorable results are highlighted in {\color{gray}gray}.}
\label{tab:ablation_results_summary_map1}
\begin{tabular}{cccccccc}
\toprule
\multicolumn{2}{c}{\textbf{Scenario}} & \multicolumn{6}{c}{\textbf{Method}} \\
\cmidrule(lr){1-2} \cmidrule(lr){3-8}
\multirow{{2}}{*}{\textbf{$v_\text{gain}$}} & \multirow{{2}}{*}{\textbf{$\omega_\text{gain}$}} & \multicolumn{2}{c}{\textbf{AFM (Ours)}} & \multicolumn{2}{c}{\textbf{PE}} & \multicolumn{2}{c}{\textbf{Online PE}} \\
\cmidrule(lr){3-4} \cmidrule(lr){5-6} \cmidrule(lr){7-8}
 & & \textbf{Success Rate} $\uparrow$ & \textbf{Steps} $\downarrow$ & \textbf{Success Rate} $\uparrow$ & \textbf{Steps} $\downarrow$ & \textbf{Success Rate} $\uparrow$ & \textbf{Steps} $\downarrow$ \\
\midrule
-1.00 & 2.00 & \textbf{1.00 $\pm$ 0.00} & \textbf{2863.40 $\pm$ 169.90} & {\color{gray}0.18 $\pm$ 0.00} & {\color{gray}4999.0 $\pm$ 0.00} & 1.00 $\pm$ 0.00 & 2879.60 $\pm$ 379.81 \\
-2.50 & 1.00 & \textbf{1.00 $\pm$ 0.00} & \textbf{3037.60 $\pm$ 442.18} & {\color{gray}0.18 $\pm$ 0.00} & {\color{gray}4999.0 $\pm$ 0.00} & 1.00 $\pm$ 0.00 & 3576.40 $\pm$ 688.12 \\
2.50 & 0.05 & {\color{gray}0.18 $\pm$ 0.00} & {\color{gray}4999.0 $\pm$ 0.00} & \textbf{0.29 $\pm$ 0.00} & {\color{gray}4999.0 $\pm$ 0.00} & {\color{gray}0.18 $\pm$ 0.00} & {\color{gray}4999.0 $\pm$ 0.00} \\
1.00 & -1.00 & \textbf{1.00 $\pm$ 0.00} & 1825.60 $\pm$ 189.14 & 1.00 $\pm$ 0.00 & \textbf{885.40 $\pm$ 27.41} & 1.00 $\pm$ 0.00 & {\color{gray}2245.40 $\pm$ 171.01} \\
1.00 & -0.50 & 0.71 $\pm$ 0.27 & 4120.60 $\pm$ 1082.86 & {\color{gray}0.53 $\pm$ 0.02} & {\color{gray}4999.0 $\pm$ 0.00} & \textbf{1.00 $\pm$ 0.00} & \textbf{2690.20 $\pm$ 260.36} \\
-1.00 & 1.00 & \textbf{1.00 $\pm$ 0.00} & 3035.80 $\pm$ 277.85 & {\color{gray}0.18 $\pm$ 0.00} & {\color{gray}4999.0 $\pm$ 0.00} & 1.00 $\pm$ 0.00 & \textbf{2815.40 $\pm$ 259.30} \\
-0.50 & 1.00 & \textbf{1.00 $\pm$ 0.00} & \textbf{4625.00 $\pm$ 327.35} & {\color{gray}0.18 $\pm$ 0.00} & {\color{gray}4999.0 $\pm$ 0.00} & 0.96 $\pm$ 0.07 & 4737.80 $\pm$ 199.80 \\
2.00 & 2.00 & \textbf{1.00 $\pm$ 0.00} & 662.80 $\pm$ 31.38 & 1.00 $\pm$ 0.00 & \textbf{492.40 $\pm$ 7.53} & 1.00 $\pm$ 0.00 & {\color{gray}675.80 $\pm$ 51.22} \\
-1.00 & -1.00 & \textbf{1.00 $\pm$ 0.00} & \textbf{3363.20 $\pm$ 345.08} & {\color{gray}0.18 $\pm$ 0.00} & {\color{gray}4999.0 $\pm$ 0.00} & 1.00 $\pm$ 0.00 & 3721.20 $\pm$ 348.69 \\
0.10 & -1.50 & {\color{gray}0.23 $\pm$ 0.03} & {\color{gray}4999.0 $\pm$ 0.00} & \textbf{0.76 $\pm$ 0.01} & {\color{gray}4999.0 $\pm$ 0.00} & 0.25 $\pm$ 0.02 & {\color{gray}4999.0 $\pm$ 0.00} \\
-0.50 & 0.50 & \textbf{0.76 $\pm$ 0.06} & {\color{gray}4999.0 $\pm$ 0.00} & {\color{gray}0.18 $\pm$ 0.00} & {\color{gray}4999.0 $\pm$ 0.00} & 0.65 $\pm$ 0.08 & {\color{gray}4999.0 $\pm$ 0.00} \\
\midrule
\bottomrule
\end{tabular}
\vspace{0.6cm}
\centering
\caption{\small Ablation Results. UGV Lifelong Dynamics Learning in Map 2 with $\delta_\mathcal{M} = 1$. Results averaged ($\pm$ standard deviation) over five seeds. Best results are shown in \textbf{bold}, while the least favorable results are highlighted in {\color{gray}gray}.}
\label{tab:ablation_results_summary_map2}
\begin{tabular}{cccccccc}
\toprule
\multicolumn{2}{c}{\textbf{Scenario}} & \multicolumn{6}{c}{\textbf{Method}} \\
\cmidrule(lr){1-2} \cmidrule(lr){3-8}
\multirow{{2}}{*}{\textbf{$v_\text{gain}$}} & \multirow{{2}}{*}{\textbf{$\omega_\text{gain}$}} & \multicolumn{2}{c}{\textbf{AFM (Ours)}} & \multicolumn{2}{c}{\textbf{PE}} & \multicolumn{2}{c}{\textbf{Online PE}} \\
\cmidrule(lr){3-4} \cmidrule(lr){5-6} \cmidrule(lr){7-8}
 & & \textbf{Success Rate} $\uparrow$ & \textbf{Steps} $\downarrow$ & \textbf{Success Rate} $\uparrow$ & \textbf{Steps} $\downarrow$ & \textbf{Success Rate} $\uparrow$ & \textbf{Steps} $\downarrow$ \\
\midrule
-1.00 & 2.00 & \textbf{1.00 $\pm$ 0.00} & \textbf{1929.00 $\pm$ 225.27} & {\color{gray}0.16 $\pm$ 0.00} & {\color{gray}4999.0 $\pm$ 0.00} & 1.00 $\pm$ 0.00 & 2012.20 $\pm$ 329.74 \\
-2.50 & 1.00 & 0.83 $\pm$ 0.34 & \textbf{3557.60 $\pm$ 744.91} & {\color{gray}0.16 $\pm$ 0.00} & {\color{gray}4999.0 $\pm$ 0.00} & \textbf{1.00 $\pm$ 0.00} & 3720.40 $\pm$ 557.46 \\
2.50 & 0.05 & {\color{gray}0.16 $\pm$ 0.00} & {\color{gray}4999.0 $\pm$ 0.00} & \textbf{0.26 $\pm$ 0.00} & {\color{gray}4999.0 $\pm$ 0.00} & 0.17 $\pm$ 0.02 & {\color{gray}4999.0 $\pm$ 0.00} \\
1.00 & -1.00 & \textbf{1.00 $\pm$ 0.00} & 1082.00 $\pm$ 96.05 & 1.00 $\pm$ 0.00 & \textbf{608.00 $\pm$ 19.04} & 1.00 $\pm$ 0.00 & {\color{gray}1388.40 $\pm$ 35.50} \\
1.00 & -0.50 & \textbf{1.00 $\pm$ 0.00} & 1981.00 $\pm$ 526.56 & {\color{gray}0.67 $\pm$ 0.17} & {\color{gray}4648.40 $\pm$ 701.20} & 1.00 $\pm$ 0.00 & \textbf{1660.60 $\pm$ 313.19} \\
-1.00 & 1.00 & \textbf{1.00 $\pm$ 0.00} & \textbf{2516.40 $\pm$ 420.09} & {\color{gray}0.16 $\pm$ 0.00} & {\color{gray}4999.0 $\pm$ 0.00} & 1.00 $\pm$ 0.00 & 2563.40 $\pm$ 525.20 \\
-0.50 & 1.00 & \textbf{1.00 $\pm$ 0.00} & \textbf{3884.60 $\pm$ 342.32} & {\color{gray}0.16 $\pm$ 0.00} & {\color{gray}4999.0 $\pm$ 0.00} & 1.00 $\pm$ 0.00 & 4025.60 $\pm$ 495.98 \\
2.00 & 2.00 & \textbf{1.00 $\pm$ 0.00} & 353.00 $\pm$ 14.18 & 1.00 $\pm$ 0.00 & \textbf{272.60 $\pm$ 2.80} & 1.00 $\pm$ 0.00 & {\color{gray}357.20 $\pm$ 14.89} \\
-1.00 & -1.00 & \textbf{1.00 $\pm$ 0.00} & 2754.00 $\pm$ 423.29 & {\color{gray}0.16 $\pm$ 0.00} & {\color{gray}4999.0 $\pm$ 0.00} & 1.00 $\pm$ 0.00 & \textbf{2241.00 $\pm$ 223.30} \\
0.10 & -1.50 & 0.19 $\pm$ 0.03 & {\color{gray}4999.0 $\pm$ 0.00} & \textbf{1.00 $\pm$ 0.00} & \textbf{4852.00 $\pm$ 38.83} & 0.20 $\pm$ 0.04 & {\color{gray}4999.0 $\pm$ 0.00} \\
-0.50 & 0.50 & \textbf{0.86 $\pm$ 0.12} & \textbf{4915.60 $\pm$ 110.71} & {\color{gray}0.16 $\pm$ 0.00} & {\color{gray}4999.0 $\pm$ 0.00} & 0.85 $\pm$ 0.08 & 4967.20 $\pm$ 63.60 \\
\midrule
\bottomrule
\end{tabular}
\end{table*}

\begin{figure*}[h!]
    \centering
    \begin{subfigure}[b]{0.49\textwidth}
        \centering
        \includegraphics[width=\textwidth]{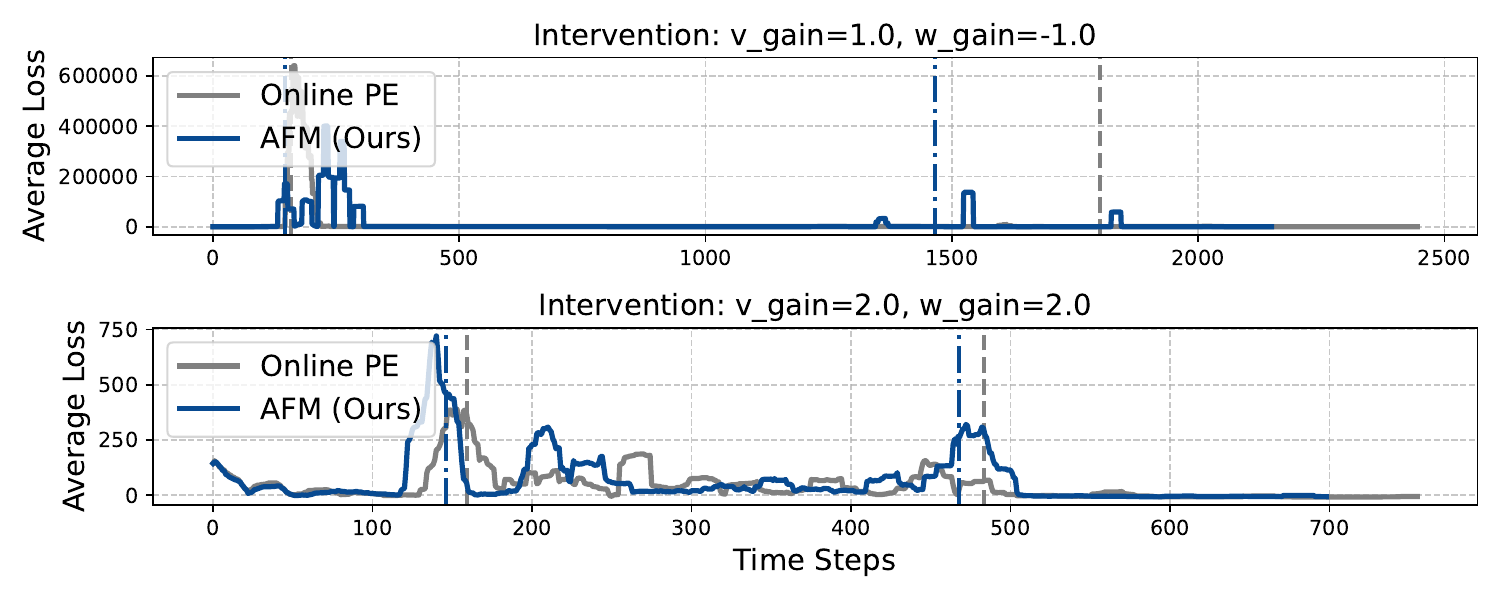}
        \caption{Map 1}
    \end{subfigure}
    \hfill
    \begin{subfigure}[b]{0.49\textwidth}
        \centering
        \includegraphics[width=\textwidth]{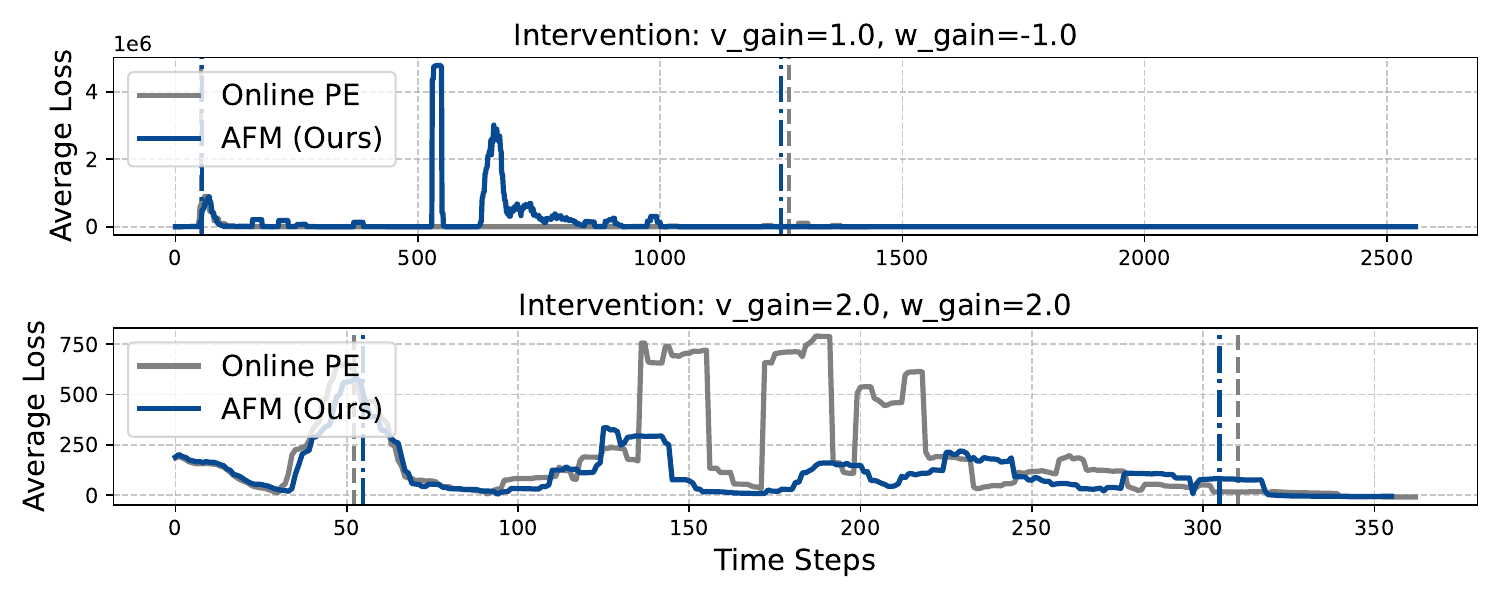}
        \caption{Map 2}
    \end{subfigure}
    \caption{\small UGV Lifelong Learning Loss Comparison. We present the results on Map 1 (left) and Map 2 (right). Vertical lines indicate the average step at which a change in the environmental dynamics occurred, thus presenting a new scenario in which the robot needs to learn to operate. These transitions coincide with the robot entering new regions of the map upon successful completion of prior objectives, thus earlier transitions are desired. For ease of visualization we only compare against the best ablation and use the moving average of the mean over five seeds with window size 20. For more details, refer to Appendix \ref{appendix:additiona_results_ugv}.}
    \label{fig:ugv_dynamics_learning_loss}
    \vspace{-3mm}
\end{figure*}

\subsection{Ablation Studies}

We conduct ablation experiments on the UGV task using two variations of the base PE dynamics model: one with frozen weights (PE) and another where the weights are updated every timestep using only the previous transition information (Online PE). The results, summarized in Tables \ref{tab:ablation_results_summary_map1} and \ref{tab:ablation_results_summary_map2}, show that in general our method (AFM) outperforms the ablations across all evaluations, with the Online PE ablation placing second. For example, in 9 out of 22 scenarios, the best-performing ablation exhibited the worst efficiency, either by exhausting the allotted time or taking longer than all other approaches, which supports AFM's ability to effectively guide lifelong dynamics learning. Furthermore, when compared with Online PE, AFM had the \textit{best} performance in \textbf{55\%} more scenarios.

Furthermore, Figure \ref{fig:ugv_dynamics_learning_loss} illustrates the average loss evolution during the task, where AFM consistently achieves lower loss and recovers to a nominal loss level after a dynamics regime change faster than the best ablation. It also reaches new dynamics regimes faster than the ablations, signaling that it took it less time to adapt to the new regimes. Additional results with more scenarios are provided in Figures \ref{fig:full_map1_loss} and \ref{fig:full_map2_loss} in Appendix \ref{appendix:additiona_results_ugv}. Figure \ref{fig:heatmaps} compares the UGV's trajectory with different continual dynamics learning methods. The plots clearly show AFM's action adjustments help the robot recover faster after an intervention.

We further investigate the potential of domain randomization (DR) to enhance the performance of AFM. To this end, we evaluate AFM in a setting where the initial dynamics model is trained with actions perturbed by $\pm$10\%, simulating inaccuracies arising from domain shift or real-world uncertainties. Our findings indicate that this AFM+DR configuration yields improved outcomes. Specifically, AFM+DR achieves an \textit{additional} \textbf{2.2\%} increase in success rate (83.6\% vs. 81.4\%) and requires, on average, \textbf{146.7} \textit{fewer steps} to reach the goal (3131.2 vs. 3277.9).

\begin{figure*}[h!] 
    \centering
    \hspace{-4mm}
    \includegraphics[width=0.24\textwidth]{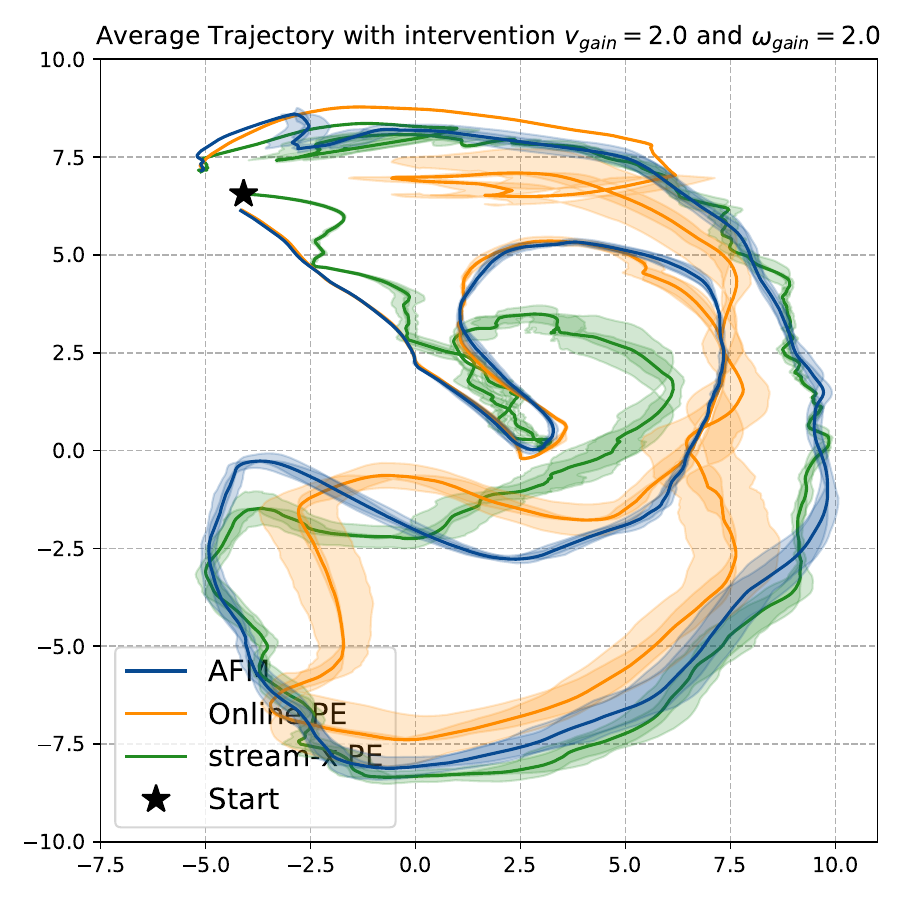}%
    \includegraphics[width=0.24\textwidth]{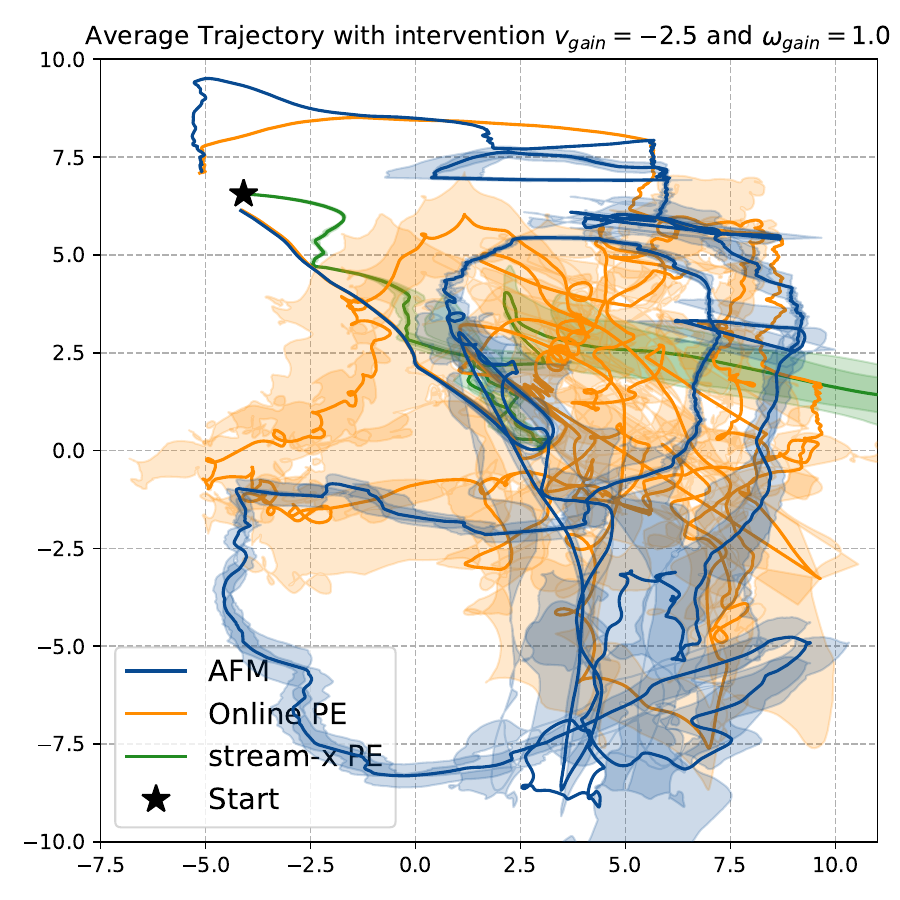}%
    \includegraphics[width=0.24\textwidth]{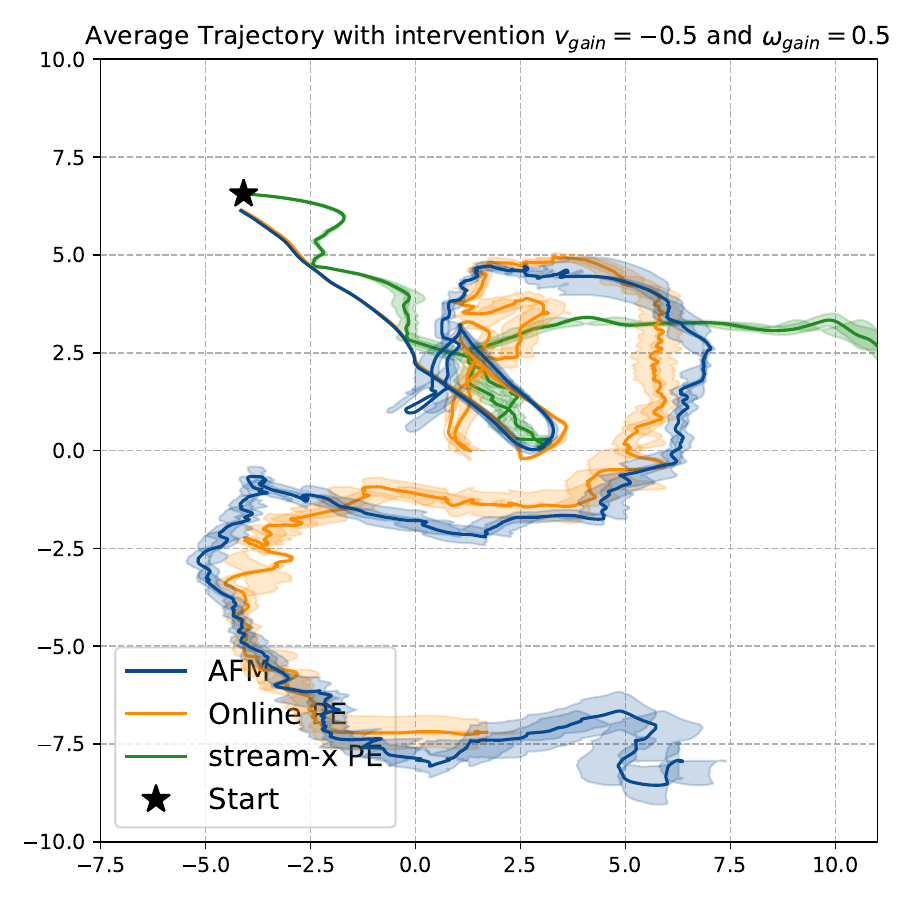}%
    \includegraphics[width=0.24\textwidth]{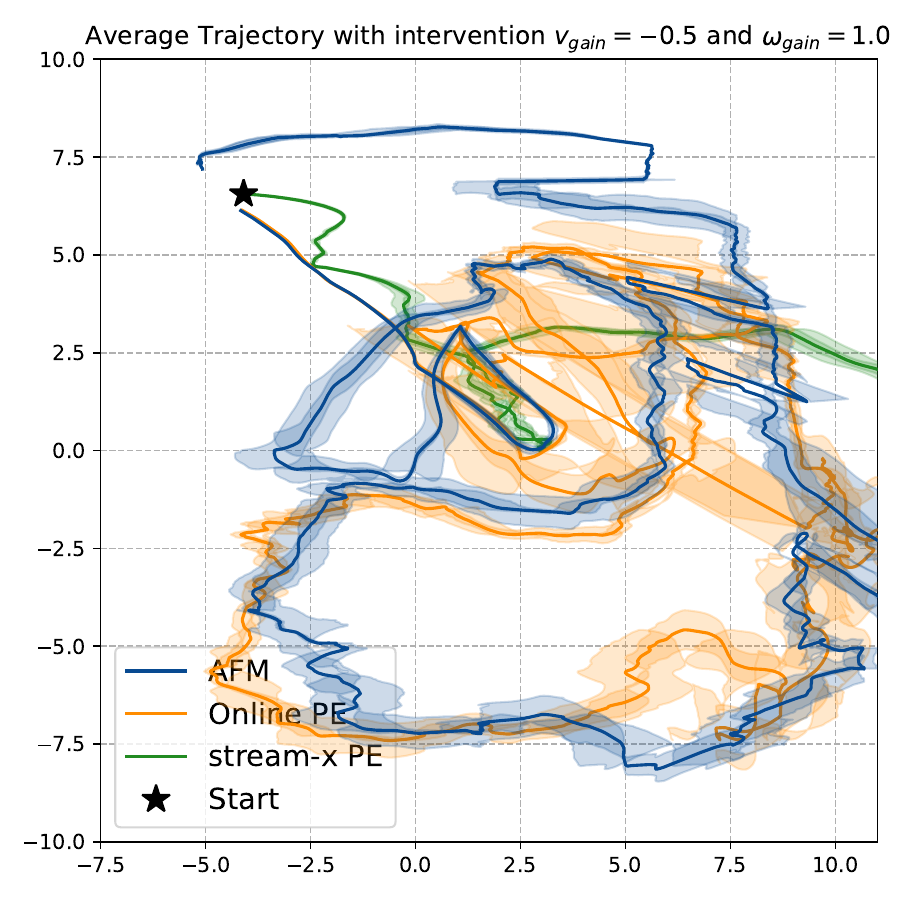}%

    \includegraphics[width=0.24\textwidth]{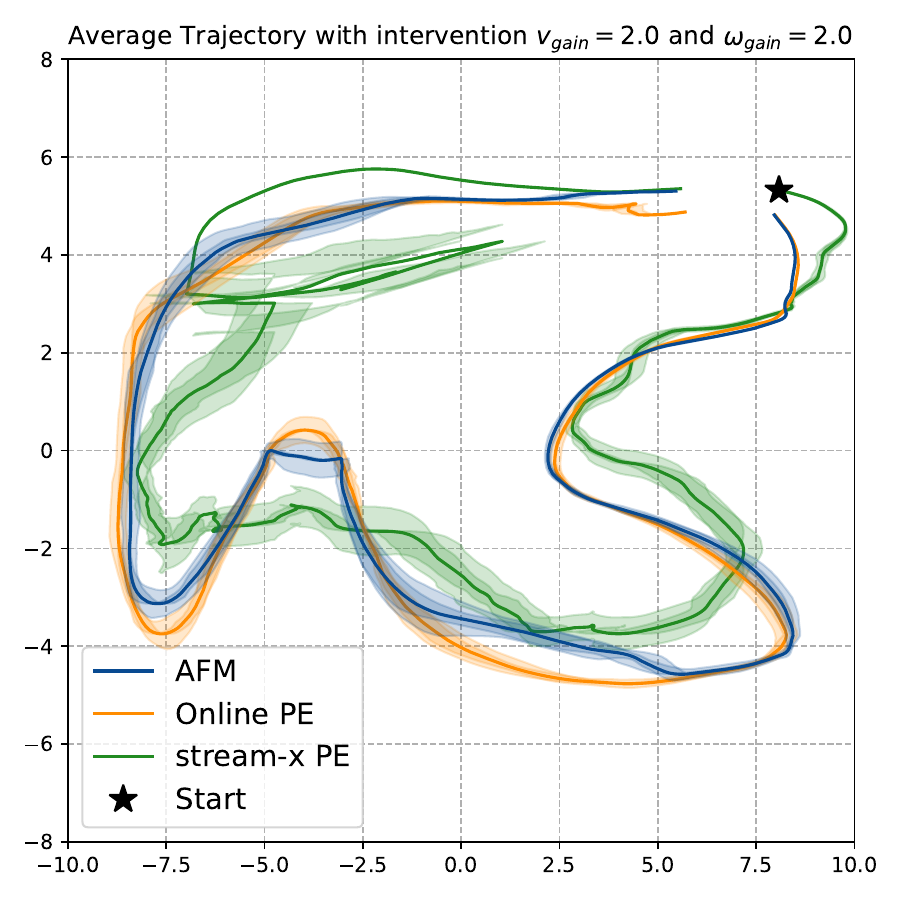}%
    \includegraphics[width=0.24\textwidth]{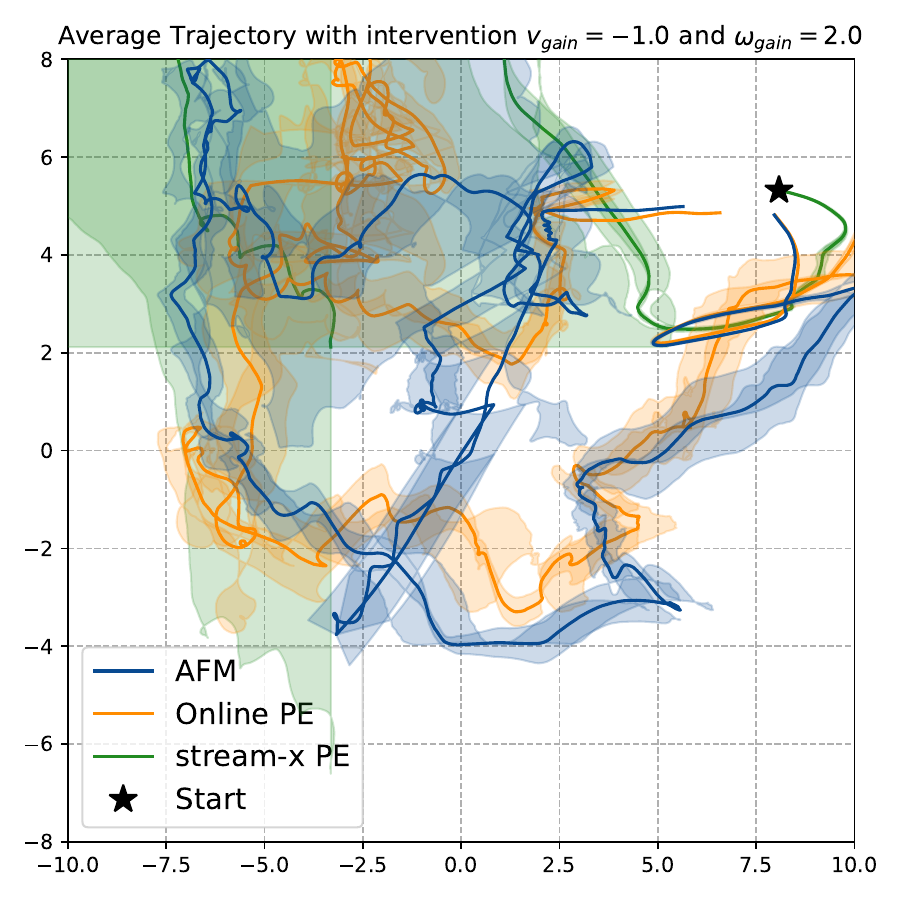}%
    \includegraphics[width=0.24\textwidth]{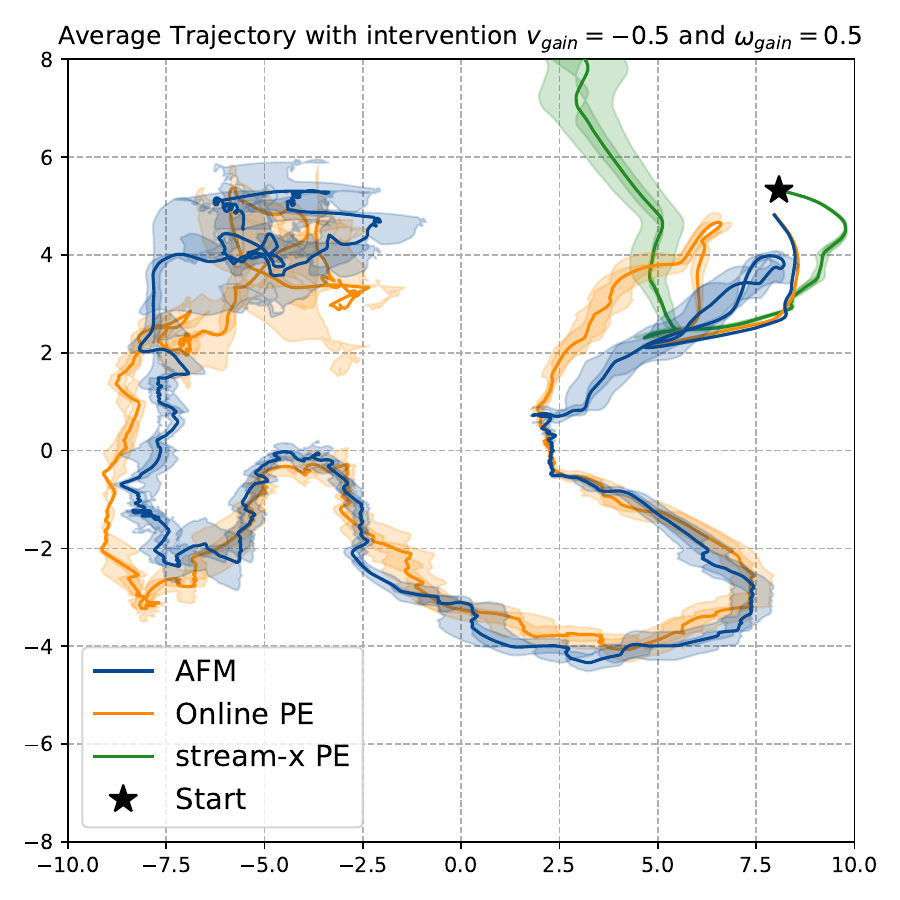}%
    \includegraphics[width=0.24\textwidth]{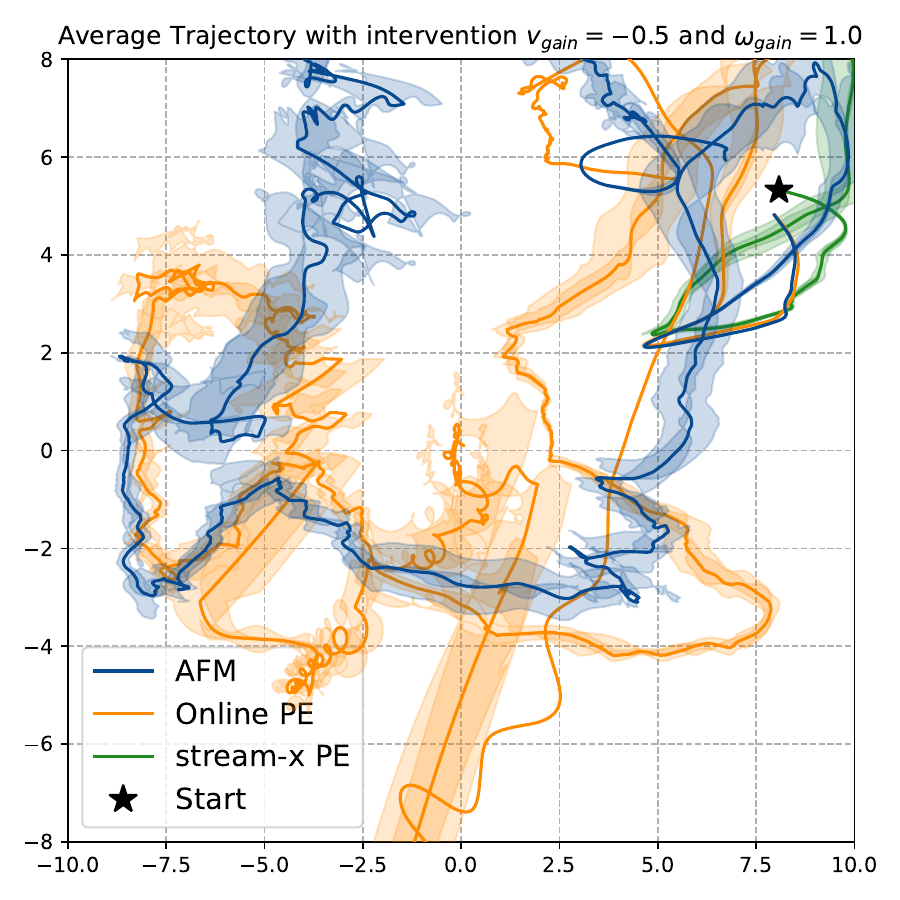}%

    \caption{Trajectory heatmaps with continual dynamics learning in Map 1 (top) and Map 2 (bottom) under multiple interventions. For clarity, we compute a moving average over 5 seeds with window size 14 and scale the standard deviation by 0.2.}
    \label{fig:heatmaps}
\end{figure*}

\section{Discussion}

We have shown that AFM offers a compelling step towards continual dynamics model learning, addressing challenges of adaptability and efficiency. We now analyze the key aspects of our findings and their implications for robotic systems.

\textit{AFM is Agnostic to the Dynamics Model Architecture:} Rather than being constrained to a specific model structure, AFM seamlessly integrates with multiple dynamics models without requiring modifications to the underlying framework. This flexibility allows practitioners to select the most suitable model for a given platform or task, rather than adapting to restrictive assumptions or architectural constraints. To demonstrate this versatility, we evaluated AFM on two distinct robotic platforms: an UGV and a quadrotor, each employing a different class of dynamics model. For the UGV, we leveraged a PE-based model, which is entirely data-driven and learns the system dynamics from experience. In contrast, the quadrotor utilized Online-KNODE-MPC, a hybrid approach that combines prior ODE-based physics models with an adaptive learning component. Despite the fundamental differences in these models---one being purely learned and the other physics-informed---AFM consistently improved performance across both cases. Specifically, for the UGV, AFM led to a 34.2\% increase in task success rate, highlighting its ability to rapidly correct for discrepancies in learned models. Meanwhile, for the quadrotor, where the best baseline model already incorporated physics-based priors and was highly specialized for the task, AFM further reduced the tracking error by 6.6\% on average. These results underscore AFM’s broad applicability in robotics, where the choice of dynamics model often varies due to platform constraints, task requirements, or data availability. By enabling effective adaptation regardless of the underlying model, AFM paves the way for more generalizable and scalable learning-based control strategies in lifelong robot autonomy.

\textit{Reconciling Exploration and Exploitation:}
The results provide strong evidence in favor of AFM as a method to balance exploration and exploitation in continual learning scenarios. Robots must navigate the tension between exploiting existing knowledge to achieve immediate goals and exploring unknown areas to adapt to a dynamics regime change and enhance long-term performance. By correcting actions to account for discrepancies in dynamics, our framework enables the robot to pursue its objectives while still uncovering useful information about its environment. Through the assessment of model misalignment, the framework determines when an acceptable alignment has been achieved, allowing the robot to fully exploit the learned model without requiring continued AFM-guided exploration. These adaptive trade-offs ensure that the robot remains productive in the short term without compromising its potential for future improvement.

\section{Conclusion and Limitations}
We have introduced AFM, a framework designed to accelerate lifelong robot dynamics model learning. AFM enables a robot to rapidly align its dynamics model, enhancing both the speed and safety of task completion. By refining the actions proposed by a model-based planner, AFM adjusts them to align more closely with those the robot would choose using a well-aligned model. This approach not only improves task performance but also facilitates more efficient learning, enabling the robot to adapt to changing environments and dynamics. The method is flexible, model-agnostic, capable of generating its own training data, and has shown significant improvements across diverse robotic platforms. This paves the way for more robust, scalable, and self-sustaining solutions for lifelong robotic autonomy.

\textit{Limitations:} Despite demonstrating significantly improved performance over competitive baselines, AFM---along with the broader challenge of continual dynamics learning---still requires further advancements. 
For example, while AFM effectively addresses the exploration phase, further improvements are needed in the generalization of the dynamics model itself. Enhancing the model’s ability to extrapolate beyond previous observations could accelerate adaptation and reduce exploration. 
Another challenge is the data-intensive nature of training FM models \cite{lipman2024flowmatchingguidecode}. While AFM mitigates this by generating high-quality training data (as described in Section \ref{sec:method}), the overall data efficiency could still be improved. 
Future algorithmic advances in this area, should translate into improved training efficiency for AFM.

\section*{Acknowledgments}
This work is supported by the National Science Foundation under Grant No. 2047169. The authors thank the anonymous reviewers for their valuable feedback, which helped improve the quality of this work.

{\fontsize{10.pt}{11.3pt}\selectfont
\bibliographystyle{plainnat}
\bibliography{references}
}
\appendix

\subsection{Streaming Model-Free Reinforcement Learning} \label{appendix:stream-x}

Streaming reinforcement learning methods aim to process data one sample at a time without storing previous samples, thereby eliminating the use of a replay buffer. This is particularly relevant for a robot that needs to incorporate the latest information into its model of the world. Traditional methods rely on a dataset of transitions which is then used to fine-tune the model. However, this is not practical nor desirable for a robot, since that would require stopping its operation while training with the whole dataset. Furthermore, additional engineering challenges would emerge, such as deciding when to stop to fine-tune or how to store all the data in a resource-constrained platform. 

To achieve streaming learning in environments with non-stationary data streams several challenges need to be addressed. For example, instability due to issues such as large per-sample gradient fluctuations, activation non-stationarity, and improper data scaling. We summarize next how the \textit{stream-x} family of algorithms addresses this \cite{elsayed2024streaming}:

\subsubsection{Sample Efficiency}
To enhance sample efficiency, the method employs sparse initialization and eligibility traces:
\begin{itemize}
    \item \textit{Sparse Initialization:} Most weights are initialized to zero while retaining a sparse set of non-zero weights, reducing interference and improving learning robustness.
    \item \textit{Eligibility Traces:} Accumulating traces facilitate credit assignment over time. The trace updates recursively using \( z_t = \gamma \lambda z_{t-1} + \nabla v(S_t, w_t) \), where \( \gamma \) and \( \lambda \) are the discount and trace parameters, respectively.
\end{itemize}

\subsubsection{Update Stability}
Step sizes are stabilized using a modified backtracking line search that bounds updates, avoiding overshooting errors in single-sample learning. An efficient step-size control mechanism mitigates computational overhead associated with traditional backtracking.

\subsubsection{Normalization and Scaling}
Layer normalization is applied across network layers to maintain stable activation distributions under non-stationary data. The normalization scale and bias parameters are not learned. Additionally, observations and rewards are scaled dynamically using unbiased mean-variance estimates to ensure effective learning despite unbounded state spaces.

These innovations collectively enable streaming deep reinforcement learning to achieve performance comparable to batch learning methods, despite the constraints of one-sample learning and non-stationary environments.

\subsection{Additional Experimental Results} \label{appendix:additiona_results_ugv}

\begin{figure*}
    \centering
    \includegraphics[width=0.5\linewidth]{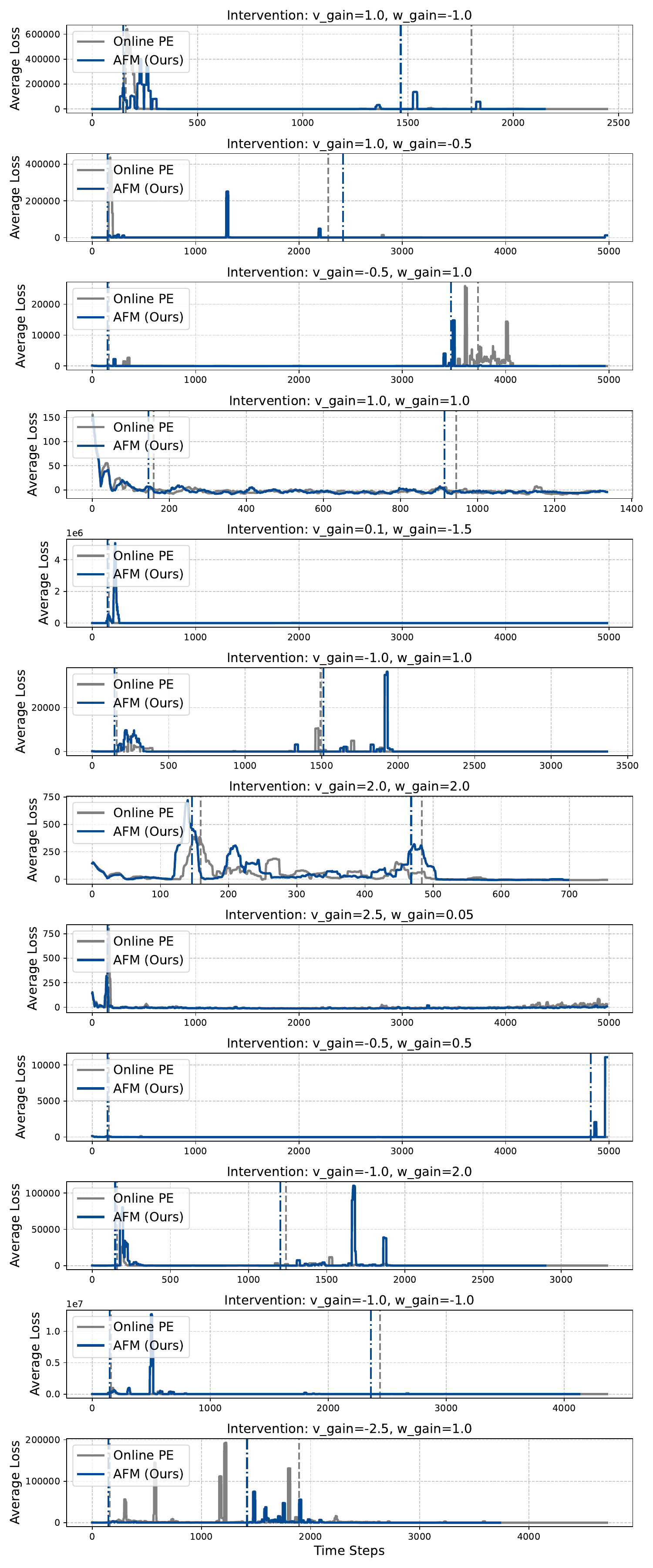}
    \caption{\small Loss Comparison for Map 1. Vertical lines indicate the average step at which a change in the environmental dynamics occurred, thus presenting a new scenario in which the robot needs to learn to operate. These transitions coincide with the robot entering new regions of the map upon successful completion of prior objectives, thus earlier transitions are desired. For ease of visualization we only compare against the best baseline and use a moving average with window size 20. We present the mean loss over five seeds.}
    \label{fig:full_map1_loss}
\end{figure*}

\begin{figure*}
    \centering
    \includegraphics[width=0.5\linewidth]{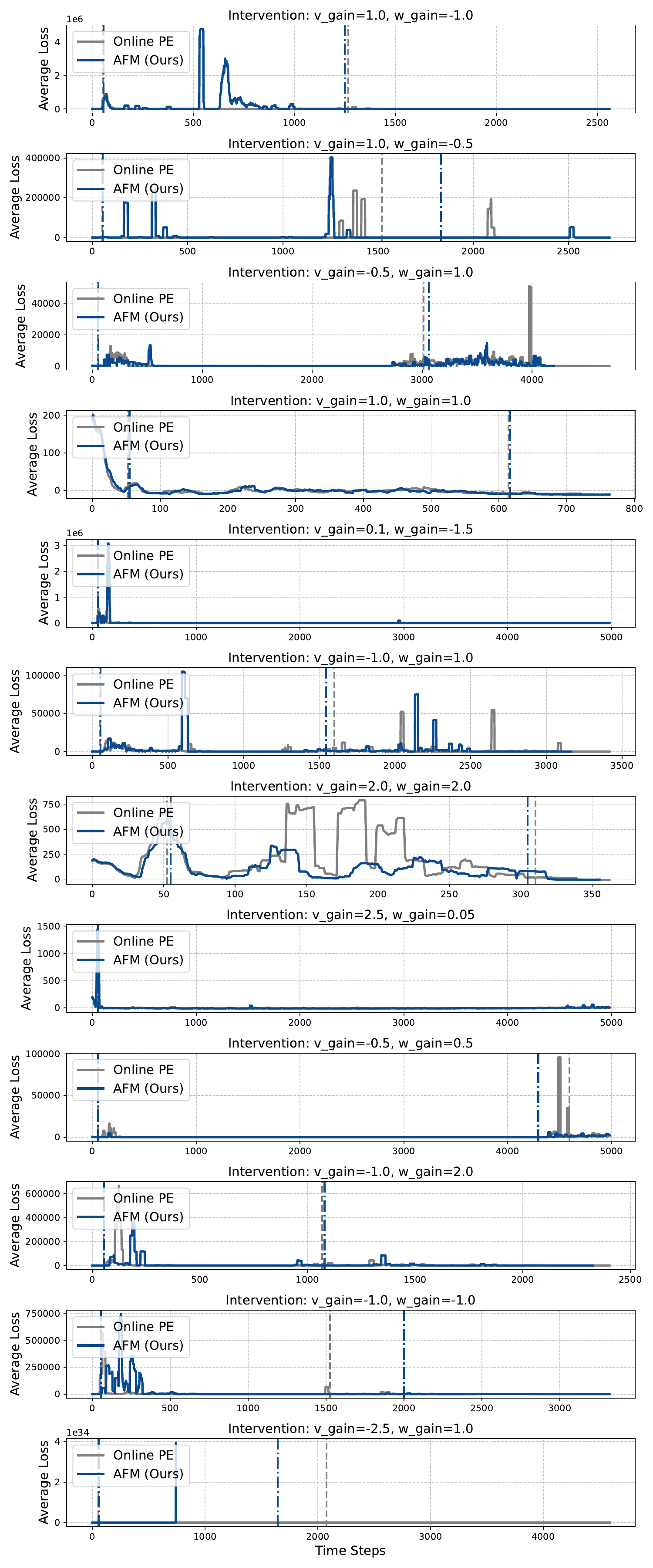}
    \caption{\small Loss Comparison for Map 2. Vertical lines indicate the average step at which a change in the environmental dynamics occurred, thus presenting a new scenario in which the robot needs to learn to operate. These transitions coincide with the robot entering new regions of the map upon successful completion of prior objectives, thus earlier transitions are desired. For ease of visualization we only compare against the best baseline and use a moving average with window size 20. We present the mean loss over five seeds.}
    \label{fig:full_map2_loss}
\end{figure*}

\textbf{Loss Comparison:} Figures \ref{fig:full_map1_loss} and \ref{fig:full_map2_loss} illustrate the loss comparison for both maps across multiple scenarios. Notably, our method consistently demonstrates accelerated learning of the environment's new dynamics. This is evident in two key observations: 1) The loss curves associated with our method generally exhibit lower values and exhibit fewer abrupt peaks, indicative of more stable and robust learning. 2) The vertical lines, signifying transitions between dynamic regimes, typically appear earlier in the curves generated by our method. This implies that the agent employing AFM reaches objectives more rapidly and transitions to subsequent goals more efficiently, a direct consequence of its accelerated learning of the underlying dynamics model.

\end{document}